\documentclass[11pt]{article}

\usepackage[final]{acl}

\usepackage{times}
\usepackage{latexsym}

\usepackage[T1]{fontenc}
\usepackage[utf8]{inputenc}

\usepackage{microtype}

\usepackage{inconsolata}

\usepackage{graphicx}

\usepackage{amsmath}
\usepackage{amssymb}
\usepackage{booktabs}
\usepackage{array}
\usepackage{multirow}
\usepackage{algorithm}
\usepackage{algpseudocode}
\usepackage{xcolor}
\usepackage[most]{tcolorbox}
\graphicspath{%
  {picture/}%
  {./}%
}
\newtcolorbox{promptbox}[1][]{%
  enhanced,
  colback=blue!3,
  colframe=blue!55!black,
  coltitle=white,
  fonttitle=\small\bfseries,
  boxrule=0.6pt,
  arc=3pt,
  left=8pt, right=8pt, top=6pt, bottom=6pt,
  boxsep=3pt,
  before upper={\setlength{\parskip}{4pt}\setlength{\parindent}{0pt}},
  title=#1,
}
\newtcolorbox{casebox}[1][]{%
  enhanced,
  colback=green!3,
  colframe=green!45!black,
  coltitle=white,
  fonttitle=\small\bfseries,
  boxrule=0.6pt,
  arc=3pt,
  left=8pt, right=8pt, top=6pt, bottom=6pt,
  boxsep=3pt,
  before upper={\setlength{\parskip}{4pt}\setlength{\parindent}{0pt}},
  title=#1,
}
\newtcolorbox{workedbox}[1][]{%
  enhanced,
  colback=orange!3,
  colframe=orange!60!black,
  coltitle=white,
  fonttitle=\small\bfseries,
  boxrule=0.6pt,
  arc=3pt,
  left=8pt, right=8pt, top=6pt, bottom=6pt,
  boxsep=3pt,
  before upper={\setlength{\parskip}{4pt}\setlength{\parindent}{0pt}},
  title=#1,
}
\newcommand{\fieldlbl}[2][gray!18]{%
  \tcbox[on line, boxrule=0pt, arc=2pt, boxsep=1pt,
         left=3pt, right=3pt, top=0.5pt, bottom=0.5pt,
         colback=#1, colframe=#1]{\footnotesize\bfseries #2}}
\title{T-Mem: Memory That Anticipates, Not Archives}

\author{
  \textbf{Weidong Guo}\textsuperscript{1*} \quad
  \textbf{Dakai Wang}\textsuperscript{1*} \quad
  \textbf{Zixuan Wang}\textsuperscript{1,2} \\
  \textbf{Hui Liu}\textsuperscript{1} \quad
  \textbf{Yu Xu}\textsuperscript{1$\dagger$} \\[2pt]
  \textsuperscript{1}Tencent \\
  \textsuperscript{2}The Hong Kong University of Science and Technology (Guangzhou) \\[2pt]
  \small{\texttt{\{weidongguo, dkkatzewang, zzixuanwang, pvopliu, henrysxu\}@tencent.com}} \\
  \small{\texttt{zwang933@connect.hkust-gz.edu.cn}} \\[2pt]
  \small{\textbf{Code and Data: }\protect\url{https://github.com/Sherlockwz/T-Mem}}
}

\usepackage{hyperref}
\begin{document}
\maketitle

% Footnote on page 1 with NO marker on the title.
% Trick: temporarily redefine \thefootnote to empty so the
% footnote symbol disappears, drop the footnotetext, then restore.
\renewcommand{\thefootnote}{}%
\footnotetext{$*$~Equal contribution.\quad $\dagger$~Corresponding author.}
% \footnotetext{Code and data: \url{https://github.com/Sherlockwz/T-Mem}.}
\renewcommand{\thefootnote}{\arabic{footnote}}

% =====================================================================
% Abstract
%   Source: abstract/abstract.md (EN, locked 2026-05-13)
% =====================================================================
\begin{abstract}
Long-term memory is essential for conversational agents to
remain coherent across extended dialogues, follow through on
commitments made many sessions earlier, and adapt their
behaviour to each user. Current LLM-backed long-term
conversational memory, however, is reachability-bounded by the
similarity between a query and stored content, both lexical and
dense-vector. The approach is effective when query and memory
share surface features such as wording or named entities (we
call this descriptive). But it misses another, equally
valuable class of cases, where query and memory do not share
surface features and are tied only by a latent semantic arc
(associative). On this regime prevailing long-term
memory systems collectively fail.
Covering this other half is what allows an assistant, for the
first time, to actively draw on past dialogue as a semantic
asset. On the memory side, this is the engineering
counterpart of what cognitive science calls episodic future
thinking: rehearsing past experience for the future contexts
under which it will need to be found.
We call these write-time rehearsals triggers.
We propose \textbf{T-Mem}, the first long-term conversational
memory architecture that covers both descriptive and
associative recall. At each of two evidence granularities,
single facts and full exchanges, T-Mem instantiates one
descriptive trigger family and one associative trigger family,
so that every memory remains reachable from both surface-similar
and relevance-bound queries. As empirical validation, T-Mem
reaches state-of-the-art on both LoCoMo and LoCoMo-Plus.
\end{abstract}

% =====================================================================
% Section 1 -- Introduction
%   Source: 1_instruction/introduction.md  (EN, locked 2026-05-13)
% =====================================================================
\section{Introduction}
\label{sec:intro}

\begin{figure}[!t]
  \centering
  \includegraphics[width=\columnwidth]{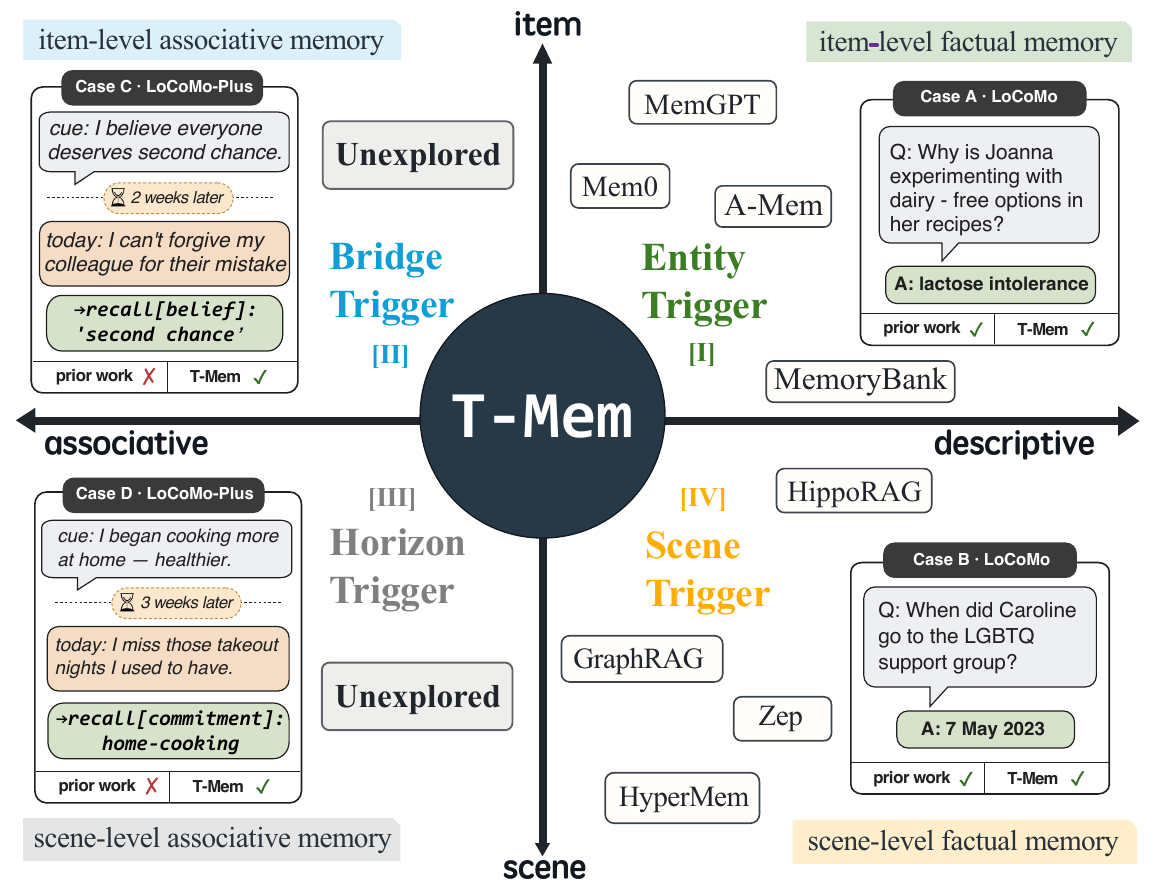}
  \caption{Trigger-design space across granularity (item/scene)
    and orientation (descriptive/associative). Prior systems
    cluster on the descriptive (right) half; T-Mem instantiates
    one trigger per quadrant.}
  \label{fig:teaser}
\end{figure}

External long-term memory modules built around large language
models let a conversational assistant reconnect each new turn
with past content: when the user later uses similar wording,
names the same entities, or anchors the same time and place, the
corresponding stored memory should be retrieved correctly. We
call this regime descriptive recall. Flat
retrieval-augmented generation \citep{Lewis2020RAG}, graph- and
hypergraph-structured memory \citep{Edge2024GraphRAG,Yue2026HyperMem},
agentic hierarchical memory
\citep{Chhikara2025Mem0,Xu2025AMEM}, and OS-style memory kernels
\citep{Packer2023MemGPT,Li2025MemOS,Wang2025MIRIX} all share the
same retrieval recipe: project query and stored content into a
single similarity space (lexical BM25 or dense embedding) and
take the nearest top-$K$. These systems differ widely in
structure but occupy the same region of the retrieval design
space (Figure~\ref{fig:teaser}, descriptive half).

% --- Figure 1 (early declaration to land at top-of-right-column on page 1) ----

Long-horizon dialogue, however, hosts another, equally common
kind of query--memory relation: query and target share no
surface form; what binds them is a latent semantic arc such as
a causal link or the continuation of a shared situation.
We call this regime associative recall, after the
cue-bound retrieval long studied in human associative memory
\citep{Anderson2014HAM}. As an example, a
month ago the user mentioned, ``a colleague on my team has a
serious seafood allergy''; today the user asks, ``where should
we take the team for dinner tonight?'' The two messages share
no surface form, yet the first is precisely what the second must
draw on. Users in long-running conversations rarely re-raise an
old topic with the same wording they used weeks earlier; they
revisit it through indirect cues triggered by a new situation
\citep{Xu2022LongTimeNoSee,Maharana2024LoCoMo}.
Similarity search, by construction, can only converge along
lexical or semantic distance; a target whose surface form has
already drifted away can never be reached from the same
neighbourhood.

Unfolding this observation gives us an orientation axis,
orientation $\in$ \{descriptive, associative\}. The axis does
not stand alone: the evidence on which a query lands also has
two natural granularities. Some queries ask about a single fact
(e.g., ``which week was Calvin's Tokyo concert?'') and are
answered at the item layer. Others ask about a coherent stretch
of dialogue (e.g., ``what else did we cover when we last
discussed Calvin's tour?'') and are answered at the scene layer.
The granularity axis applies on both sides of orientation and
is therefore orthogonal to it, opening a $2\!\times\!2$
retrieval space (Figure~\ref{fig:teaser}). Prevailing memory
systems concentrate in Quadrant~I (item~$\times$~descriptive),
with recent graph-based systems extending toward Quadrant~IV
(scene~$\times$~descriptive); Quadrants~II and~III, the
associative half, remain a structural blind spot.

We propose \textbf{T-Mem}, the first long-term conversational
memory architecture that covers all four quadrants. T-Mem
organises memory at two evidence granularities, scene (a
coherent stretch of dialogue) and item (an atomic fact extracted
from a scene). On top of this evidence layer, T-Mem instantiates
one trigger family per quadrant --- Entity (Q~I), Bridge
(Q~II), Horizon (Q~III), and Scene (Q~IV); Bridge and Horizon
are precisely the two families that fill in the associative
blind spot identified above. All four
families are precomputed offline by a memory-construction LLM at
write time and stored alongside their host item or scene. This
decouples what makes a memory reachable from what counts as
evidence at answer time: when a query no longer reaches the host
through similarity, it can still locate the host via one of its
triggers. As a result, every memory
remains reachable both from descriptively similar and from
associatively relevant queries.

Our contributions are summarised as follows:

\noindent (i)~We show that prevailing long-term
conversational memory systems share a single similarity-based
retrieval recipe and therefore cover only the descriptive half
of a granularity~$\times$~orientation design space
(Q~I and Q~IV); associative recall (Q~II and Q~III) is a
structural blind spot of this recipe.

\noindent (ii)~We propose T-Mem, a long-term
conversational memory architecture that closes this blind spot
at the indexing layer: on top of a scene--item evidence layer,
T-Mem instantiates one write-time trigger family per quadrant of
the design space, bringing all four quadrants under one system.

\noindent (iii)~Empirically, T-Mem reaches state-of-the-art on
both LoCoMo (80.26\%) and LoCoMo-Plus (74.81\%), and tightens
the cross-benchmark gap from the 28--50\,pp range of prevailing
systems to 5.45\,pp.

% =====================================================================
% Section 2 -- Related Work
%   Source: 2_relatedwork/relatedwork.md (EN, locked 2026-05-13)
%   ZH parity: 2_relatedwork/relatedwork_zh.md
%   Cite count: 18 (16 reused from §1 + 2 new psychology entries
%   appended to custom.bib in the same commit:
%     Suddendorf2007MentalTimeTravel,
%     AddisSchacter2008ConstructiveSimulation).
% =====================================================================
\section{Related Work}
\label{sec:related}

Long-term conversational memory has converged on four broad design
families \citep{Zhang2024MemorySurvey}. We organise
prior work along the two axes used in Section~\ref{sec:intro}:
granularity (item vs.\ scene) and orientation
(descriptive vs.\ associative).

\subsection{Long-Term Memory Architectures}
\label{sec:rw_arch}

\noindent\textbf{(i) Flat RAG.}~Retrieval-augmented generation
\citep{Lewis2020RAG} chunks and dense-indexes the dialogue stream
and retrieves by query embedding; it is the baseline on which all
later families build.

\noindent\textbf{(ii) Graph-, hypergraph-, and temporal-graph
structured memory.}~A second family adds explicit structure:
knowledge graphs for query-focused summarisation (GraphRAG
\citealp{Edge2024GraphRAG} and variants
\citealp{Guo2025LightRAG, Gutierrez2025HippoRAG}), temporal graphs
for agent memory (Zep \citealp{Rasmussen2025Zep}), and hyperedges
over a topic--episode--fact hierarchy (HyperMem
\citealp{Yue2026HyperMem}).

\noindent\textbf{(iii) Agentic hierarchical memory.}~A third family
condenses each session into fine-grained notes and consolidates
them incrementally: forgetting-curve user facts (MemoryBank
\citealp{Zhong2024MemoryBank}), production-oriented fact layers
(Mem0 \citealp{Chhikara2025Mem0}), dynamic re-linking across notes
(A-MEM \citealp{Xu2025AMEM}), with lightweight
\citep{Fang2025LightMem} and autonomous-augmentation
\citep{Salama2025MemInsight,Huang2026AMA} variants.

\noindent\textbf{(iv) OS-style memory kernels.}~A fourth family
frames memory as hierarchical
storage with explicit scheduling and paging policies, with MemGPT
\citep{Packer2023MemGPT}, MemOS \citep{Li2025MemOS}, and MIRIX
\citep{Wang2025MIRIX} as representatives.

Together these families occupy Quadrants~I and~IV of
Figure~\ref{fig:teaser}: item-only systems (flat RAG, agentic
stacks, most OS-kernels) sit in~I, structured memory extends
toward~IV. Quadrants~II and~III remain systematically
under-served.

\subsection{Cognitive-Cue Memory Access}
\label{sec:rw_cue}

A separate line draws on classical associative memory
\citep{Anderson2014HAM}, where recall is cue-driven: a current
utterance activates a temporally distant episode through learned
associations rather than surface similarity. This aligns with
what cognitive science calls episodic future thinking
\citep{Suddendorf2007MentalTimeTravel,AddisSchacter2008ConstructiveSimulation},
in which past experience is rehearsed for future cues. LoCoMo
\citep{Maharana2024LoCoMo} and LoCoMo-Plus
\citep{Li2026LoCoMoPlus} operationalise this future-oriented view
on the evaluation side; T-Mem instantiates it on the retrieval
side, with the Bridge and Horizon Triggers occupying
Quadrants~II and~III respectively.

\subsection{Persona and Profile Memory}
\label{sec:rw_persona}

A complementary body of work treats per-speaker profiles as a
first-class ingredient of long-form dialogue. Persona-grounded
benchmarks
\citep{Zhang2018Personalizing,Mazare2018PersonalizedAgents}
established that speakers carry structured attributes retrievable
across turns, and long-horizon variants
\citep{Xu2022GoldfishMemory,Xu2022LongTimeNoSee,Jang2023ConversationChronicles}
argue that such profiles should aggregate across distant sessions.
MemoryBank \citep{Zhong2024MemoryBank} and Mem0
\citep{Chhikara2025Mem0} instantiate this view inside agentic
memory systems. We treat this functionality as
complementary, not as a substitute for fine-grained recall:
T-Mem includes a per-speaker Persona that fills the
profile-coverage gap left by sparse trigger output.

% =====================================================================
% Section 3 -- Approach
%   Source: 3_approach/approach.md (EN, locked 2026-05-13)
% =====================================================================
% --- Figure 2 (double-column) ------------------------------------
\begin{figure*}[!t]
  \centering
  \includegraphics[width=\textwidth]{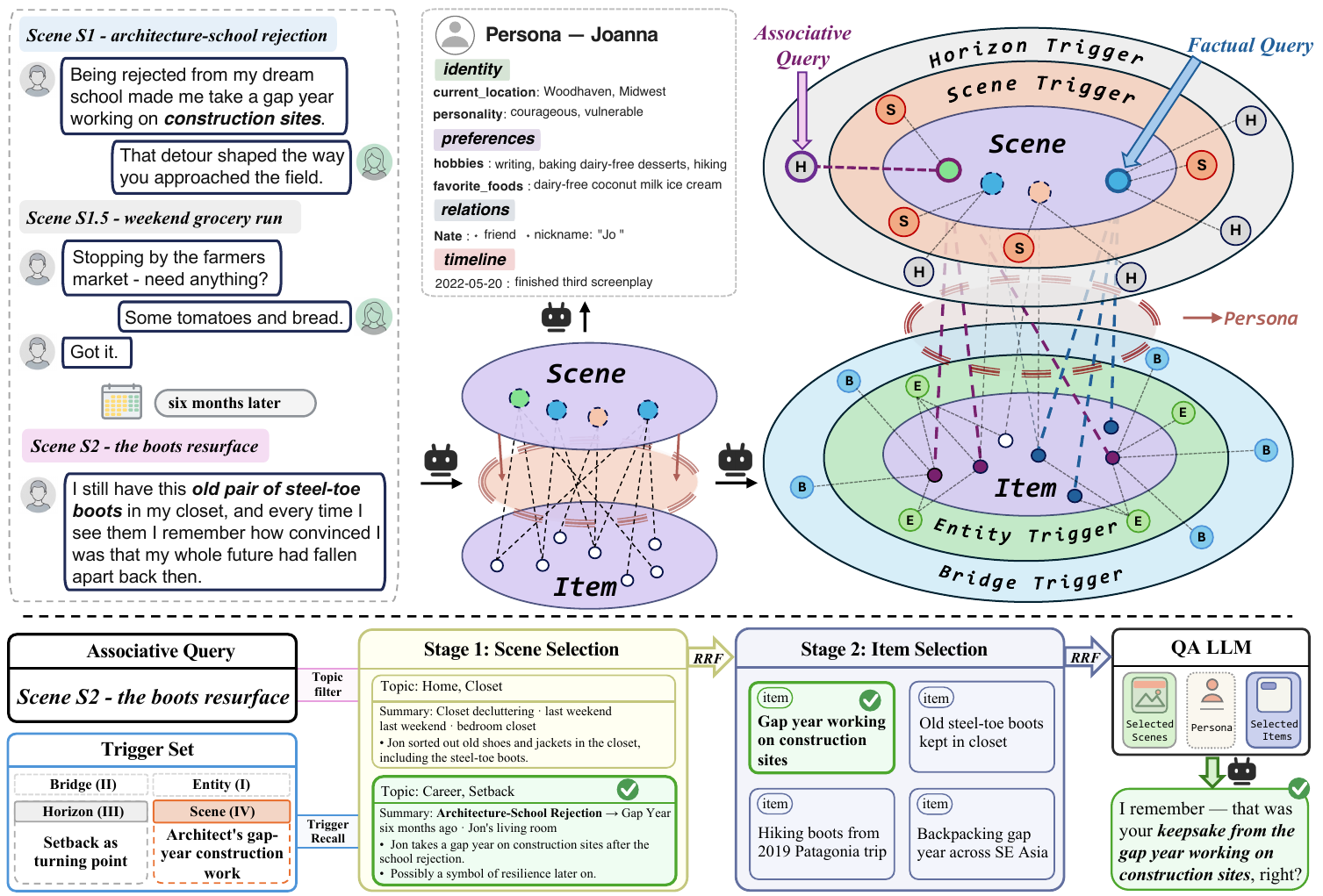}
  \caption{Framework of T-Mem. The construction pipeline
    segments scenes, assigns topics, extracts items, and
    instantiates one trigger family per quadrant (colours of
    scene nodes indicate topic membership). The retrieval
    cascade performs top-down topic\,$\to$\,scene\,$\to$\,item
    selection with associative-trigger augmentation.}
  \label{fig:tmem-arch}
\end{figure*}

\section{Approach}
\label{sec:approach}

We design T-Mem as a memory architecture organised by retrieval
capability rather than by storage form, with each component
allocated to one quadrant of the cue--memory space of
\S\ref{sec:intro}. Figure~\ref{fig:tmem-arch} gives an overview;
the section proceeds in three passes:
a typed memory structure, a four-stage construction pipeline,
and a top-down retrieval cascade.

% --- Table 1 (worked example, single-column) --------------------
\begin{table}[t]
  \centering
  \small
  \setlength{\tabcolsep}{4pt}
  \renewcommand{\arraystretch}{1.25}
  \begin{tabular}{@{}p{0.20\columnwidth}p{0.72\columnwidth}@{}}
    \toprule
    \textbf{Field} & \textbf{Worked example} \\
    \midrule
    Topic ($v^{T}$) &
      Gina's Clothing Store Ad Campaign Launch. \\
    Scenes ($v^{S}_{1}, v^{S}_{2}$) &
      \textbf{$v^{S}_{1}$, 29\,Jan} --- Gina launches her
      clothing-store ad campaign and Jon congratulates her.\newline
      \textbf{$v^{S}_{2}$, 1\,Feb} --- Gina secures a wholesaler
      and shows Jon her newly designed store interior. \\
    Item ($v^{I}$) &
      On 1\,Feb\,2023, Gina expanded her clothing store after a
      wholesaler responded positively to her outreach.
      (spans $v^{S}_{1}$ + $v^{S}_{2}$) \\
    Persona ($\mathcal{X}$) &
      \texttt{identity.occupation} = ``fashion intern'';
      \texttt{preferences.hobbies} $\supseteq$ \{``running a
      clothing store'', ``contemporary dance''\}. \\
    \bottomrule
  \end{tabular}
  \caption{A worked example: two scenes on different days are
    grouped under one topic, anchoring an item --- Gina's store
    expansion --- that depends on both. The Persona row aggregates
    speaker attributes that no single item carries on its own.}
  \label{tab:worked-example}
\end{table}

\subsection{Memory Structure}
\label{sec:approach-structure}

T-Mem operates over five kinds of objects (Table~\ref{tab:worked-example} gives a worked instance). We introduce them
one by one before giving a unified summary.
\begin{itemize}\itemsep0pt\parskip2pt
\item \textbf{scenes} ($\mathcal{V}^{S}$): cohesive exchanges
  segmented from the dialogue stream; handed to the QA LLM as
  evidence at answer time.
\item \textbf{items} ($\mathcal{V}^{I}$): atomic facts extracted
  from scenes, anchored via $\mathcal{E}^{SI}$ to one or more
  host scenes; also handed to the QA LLM as evidence. An
  atomic item attaches to a single source scene, while a
  connected item attaches to every source scene it draws
  from.
\item \textbf{topic labels} ($\mathcal{V}^{T}$): a multi-label
  tagging produced by a lightweight topic module, connected to
  scenes via $\mathcal{E}^{TS}$. Topic labels are used
  only to scope item extraction and to pre-filter
  the scene pool at retrieval; they are never read
  by the QA LLM.
\item \textbf{four trigger families}
  ($\mathcal{T}^{\mathrm{Ent}}, \mathcal{T}^{\mathrm{Brg}}$ on
  items; $\mathcal{T}^{\mathrm{Scn}}, \mathcal{T}^{\mathrm{Hor}}$
  on scenes): one per quadrant of the design space of
  \S\ref{sec:intro}; they participate in retrieval only.
\item \textbf{Persona} ($\mathcal{X}$): a per-speaker summary of
  standing traits, aggregated across the full conversation. It
  is injected as ambient context at answer time outside the
  retrieval channel, appended after the retrieved scenes and
  items rather than competing for the retrieval budget.
\end{itemize}

Putting these five kinds of objects together, T-Mem stores
conversational memory as the typed tuple
\begin{equation}
  \label{eq:memory-tuple}
  \begin{aligned}
    \mathcal{M} \;=\; \bigl(\,
      &\mathcal{V}^{T}\!\cup\!\mathcal{V}^{S}\!\cup\!\mathcal{V}^{I},\;\;
       \mathcal{E}^{TS}\!\cup\!\mathcal{E}^{SI}, \\
      &\mathcal{T}^{\mathrm{Ent}}\!\cup\!\mathcal{T}^{\mathrm{Brg}}
       \!\cup\!\mathcal{T}^{\mathrm{Scn}}\!\cup\!\mathcal{T}^{\mathrm{Hor}},\;\;
       \mathcal{X}\,\bigr).
  \end{aligned}
\end{equation}
Three design commitments shape this object: evidence
layers are kept type-segregated so that scenes and items can
each be retrieved at their own granularity; topic labels are
kept off the QA channel so that pre-filtering does not
contaminate evidence; and triggers are kept off the evidence
path so that ``how a memory is reached'' is decoupled from
``what is reached''.

\subsection{Memory Construction}
\label{sec:approach-construction}

$\mathcal{M}$ is produced by a four-stage pipeline:
scene segmentation, topic assignment, item extraction, and
trigger instantiation.
The order is load-bearing, since each stage materialises a
structural commitment that no later stage can otherwise recover.

\subsubsection{Scenes and Topics}
\label{sec:approach-construction-scenes}

Session boundaries in long-term dialogue are an artefact of data
collection rather than event closure: a session can mix several
events, and one event can span several sessions.
T-Mem therefore segments scenes by event closure, using a
lightweight boundary-detector LLM that scans a sliding turn buffer
and emits a scene whenever the current event closes.

Each scene node carries four fields with three distinct
read-paths:
\begin{itemize}\itemsep0pt\parskip2pt
\item title and summary --- consumed by the
  lexical and dense indices;
\item raw turn sequence --- the evidence handed to the QA
  LLM at answer time;
\item third-person narrative --- the input read by the
  Scene and Horizon trigger extractors.
\end{itemize}

Scenes that belong to the same recurring subject can be scattered
far apart along the dialogue stream, leaving an extractor that
sees one scene at a time no way to recover their cross-scene
logical links. T-Mem therefore processes scenes in arrival order:
for each new scene, a lightweight topic module decides which
existing topic labels it should be admitted to, and opens a new
label whenever none fits. Topic labels grow from the data, and a
scene can be admitted to several at once.

\subsubsection{Item Extraction}
\label{sec:approach-construction-items}

Conversational queries split into two classes: those that
interrogate a fact in isolation, and those that interrogate the
relation between two facts. A single granularity of items cannot
serve both. Queries of the first kind are best served by a
unit that distills the fact away from its scene. Queries of
the second kind are best served by a unit that preserves the
cross-scene logical link before retrieval ever sees it. For
every topic label
$v^{T}\!\in\!\mathcal{V}^{T}$ we therefore invoke the extractor
LLM once and obtain two complementary item types in the same
response. An atomic item is linked via $\mathcal{E}^{SI}$
to its single source scene, while a connected item is
linked via $\mathcal{E}^{SI}$ to every source scene it
draws from.

\subsubsection{Trigger Instantiation}
\label{sec:approach-construction-triggers}

The four trigger families do not play equivalent roles: Entity
and Scene serve the descriptive half of the design space (already
covered by similarity search), while Bridge and Horizon serve
the associative half (unreachable by similarity search and
identified as the blind spot in \S\ref{sec:intro}).

\textbf{Entity Trigger (Q\,I).} Names the item with a
superordinate concept. Entity and Bridge are produced jointly
within a single prompt call (one $N$-trigger generation per
item), not in separate stages.

\textbf{Bridge Trigger (Q\,II).} Projects the item onto a
situation in which knowing this item would matter, even when
the surface form is far away (e.g., an item about ``an allergy''
projects to choosing a restaurant for a team dinner). Each
trigger comes with a one-clause rationale. No public benchmark
yet isolates item-level associative recall, so Bridge's
end-to-end contribution is architecturally instantiated but not
directly measurable in \S\ref{sec:exp} (see Limitations).

\textbf{Scene Trigger (Q\,IV).} Describes the current scene
along four orthogonal attributes (situation, object, event, and
emotion), one sentence each.

\textbf{Horizon Trigger (Q\,III).} Projects the same scene onto
a set of forward-looking dimensions, so that the scene can still
be reached when a future query approaches it from a related but
different situation.

\subsection{Indexing}
\label{sec:approach-indexing}

The two index types serve the two retrieval axes of
\S\ref{sec:intro}: Node Indices support the descriptive
axis along which a query restates a host scene or item, and
Trigger-aware Indices support the associative axis along
which a query reaches a host through a learned link.

\paragraph{Node Indices.}
Every node is registered into a shared BM25 corpus and a
per-type dense table.

\paragraph{Trigger-aware Indices.}
Each trigger family surfaces host nodes on behalf of the
query, not the triggers themselves. Each item exposes three views
(concept-only, bridge-only, joint $=$
concept\,$\Vert$\,bridge\,$\Vert$\,rationale), independently
encoded; the cosine score is the nan-aware max across views and is
attributed to the host item. Scene-level triggers follow the
analogous multi-view scheme over scene attributes and Horizon
channels.

% --- Algorithm 1 (online retrieval sketch, single-column) ---------
\begin{algorithm}[t]
\small
\caption{T-Mem online retrieval (sketch).}
\label{alg:retrieval-main}
\begin{algorithmic}[1]
\Require query $q$;\, top-$K$ budgets $k^{\mathrm{T}}\!,k^{\mathrm{S}}\!,k^{\mathrm{I}}$
\Ensure  scenes $\mathcal{R}_{S}$,\, items $\mathcal{R}_{I}$,\, persona $X$
\State $\mathcal{R}_{T}\gets\mathrm{top}_{k^{\mathrm{T}}}\,\mathrm{RRF}\bigl(q,\mathcal{V}^{T}\bigr)$\hfill\Comment{topic prefilter}
\State $\mathcal{C}_{2}\gets\mathcal{E}^{TS}(\mathcal{R}_{T})\cup\mathrm{SceneCue}(q)$\hfill\Comment{$+$ Scn, Hor}
\State $\mathcal{R}_{S}\gets\mathrm{top}_{k^{\mathrm{S}}}\,\mathrm{RRF}\bigl(q,\mathcal{C}_{2}\bigr)$
\State $\mathcal{C}_{3}\gets\mathcal{E}^{SI}(\mathcal{R}_{S})\cup\mathrm{TrigRecall}(q;\tau)$\hfill\Comment{$+$ Ent, Brg}
\State $\mathcal{R}_{I}\gets\mathrm{top}_{k^{\mathrm{I}}}\,\mathrm{RRF}\bigl(q,\mathcal{C}_{3}\bigr)$
\State $X\gets\mathrm{Persona}\bigl(\mathrm{speaker}(q)\bigr)$\hfill\Comment{ambient}
\State \Return $(\mathcal{R}_{S},\,\mathcal{R}_{I},\,X)$
\end{algorithmic}
\end{algorithm}

% --- Table 2 (LoCoMo main results) -------------------------------
% LoCoMo main table.  Layout: 9 main rows (matched QA pipeline, GPT-4o-mini
% + official LoCoMo QA prompt) + 2 reference-comparison rows (HyperMem own
% QA pipeline, GPT-4.1-mini + 7-step CoT).  Numbers locked from _results.md.
\begin{table*}[t]
  \centering
  \small
  \setlength{\tabcolsep}{4pt}
  \begin{tabular}{l c c c c c c}
    \toprule
    \textbf{Method} & \textbf{Single-hop} & \textbf{Multi-hop} & \textbf{Temporal} & \textbf{Open Domain} & \textbf{Overall} & \textbf{F1} \\
    \midrule
    MIRIX                                            & 68.22 & 54.26 & 68.54 & 46.88 & 64.33 & 28.10 \\
    Mem0                                             & 73.33 & 58.75 & 52.34 & 45.83 & 64.57 & 43.46 \\
    Zep                                              & 66.23 & 52.12 & 54.82 & 33.33 & 59.22 & 41.23 \\
    Memobase                                          & 73.12 & 64.65 & 81.20 & 53.12 & 72.01 & 50.18 \\
    MemU                                              & 66.34 & 63.12 & 27.10 & 50.00 & 56.55 & 35.15 \\
    Supermemory                                       & 67.30 & 51.12 & 31.77 & 42.67 & 55.34 & 34.87 \\
    MemOS                                            & 81.09 & 67.49 & 75.18 & \textbf{55.90} & 75.80 & 45.27 \\
    HyperMem                                          & 84.34 & 62.29 & 79.44 & 47.92 & 77.01 & 49.93 \\
    \textbf{T-Mem (Ours)}                             & \textbf{85.97} & \textbf{69.15} & \textbf{82.55} & 55.21 & \textbf{80.26} & \textbf{51.96} \\
    \midrule
    HyperMem (reported)$^{\dagger}$                              & 96.08 & 93.62 & 89.72 & 70.83 & 92.73 & 15.78 \\
    \textbf{T-Mem (matched)}$^{\dagger}$                          & \textbf{97.15} & 93.62 & \textbf{91.28} & \textbf{71.88} & \textbf{93.70} & \textbf{16.97} \\
    \bottomrule
  \end{tabular}
  \caption{LoCoMo results (LLM-as-judge accuracy, \%). Rows
    marked with $\dagger$ reuse HyperMem's own QA pipeline,
    which deviates from the official LoCoMo pipeline
    (see Appendix~\ref{sec:appendix-repro}); all other rows
    follow the official pipeline, with numbers for MIRIX, Mem0,
    Zep, Memobase, MemU, Supermemory and MemOS taken from
    \citet{Li2025MemOS}.}
  \label{tab:locomo}
\end{table*}

% --- Table 3 (LoCoMo-Plus main results) --------------------------
\begin{table}[t]
  \centering
  \small
  \setlength{\tabcolsep}{4pt}
  \begin{tabular}{l c c c}
    \toprule
    \textbf{Method} & \textbf{LoCoMo} & \textbf{LoCoMo-Plus} & \textbf{Gap} \\
    \midrule
    GPT-4o                                & 66.14 & 21.05 & 45.09 \\
    Gemini-2.5-Pro                        & 71.46 & 26.06 & 45.40 \\
    \midrule
    Mem0                                  & 64.94 & 15.80 & 49.14 \\
    SeCom                                 & 64.97 & 14.90 & 50.07 \\
    A-Mem                                 & 66.70 & 17.20 & 49.50 \\
    MemOS                                 & 75.80 & 32.67 & 43.13 \\
    HyperMem                              & 77.01 & 48.63 & 28.38 \\
    \textbf{T-Mem (Ours)}                 & \textbf{80.26} & \textbf{74.81} & \textbf{5.45} \\
    \bottomrule
  \end{tabular}
  \caption{LoCoMo vs.\ LoCoMo-Plus (LLM-as-judge accuracy, \%).
    \textbf{Gap} is the drop from LoCoMo to LoCoMo-Plus.
    Numbers for Mem0, SeCom, A-Mem, GPT-4o and Gemini-2.5-Pro
    are taken from \citet{Li2026LoCoMoPlus}; the adversarial
    category is excluded following common LoCoMo practice.
    GPT-4o and Gemini-2.5-Pro are evaluated zero-shot on the
    complete raw dialogue history, with no retrieval or memory
    mechanism.}
  \label{tab:locomoplus}
\end{table}

\subsection{Retrieval}
\label{sec:approach-retrieval}

Given a query, T-Mem feeds the QA LLM through a top-down
topic\,$\to$\,scene\,$\to$\,item cascade, scoring candidates at
each layer with reciprocal rank fusion (RRF) over the lexical and
dense rankings of \S\ref{sec:approach-indexing}:
\begin{equation}
  \label{eq:rrf}
  \mathrm{RRF}(d) \;=\;
  \sum_{m=1}^{M}
  \frac{1}{k_{0} \;+\; \mathrm{rank}_m(d)},
\end{equation}
where $M$ is the per-call number of fused ranklists and $k_{0}$
is the smoothing constant.
Algorithm~\ref{alg:retrieval-main} sketches the per-layer flow.

\paragraph{Cascade ordering and trigger bypass.}
The top-down ordering is the structural invariant of the
cascade: each upper layer is coarser by construction, so pushing
the cheapest cut to the front lets the scene and item layers each
operate on a small fraction of the corpus. Crucially, scenes and items reached through any of
the four triggers are not gated by the topic prefilter
(Stage~1 of Algorithm~\ref{alg:retrieval-main}). The prefilter is,
by construction, a similarity-based neighbourhood test
(BM25\,+\,dense over topic labels), while the Bridge and Horizon
Triggers (\S\ref{sec:approach-construction-triggers}) are designed
precisely to surface on cues that fall outside any such
neighbourhood. Gating them by surviving topics would re-impose
the same similarity-only retrieval regime that
\S\ref{sec:intro} sets T-Mem against.

% =====================================================================
% Section 4 -- Experiments
%   Source: 4_experiment/experiment.md + _results.md (EN, 2026-05-12)
% =====================================================================
\section{Experiments}
\label{sec:exp}

\subsection{Experimental Setup}
\label{sec:exp-benchmarks}

\paragraph{Benchmarks.}
We evaluate on two long-term-memory benchmarks. \textbf{LoCoMo}
\citep{Maharana2024LoCoMo} comprises multi-session conversations
spanning weeks to months, with four question types:
Single-hop (recover a single fact stated once in the
dialogue), Multi-hop (combine facts spanning two or more
sessions), Temporal (resolve when something happened), and
Open Domain (answer using a speaker's persona traits or
world knowledge beyond the dialogue text). \textbf{LoCoMo-Plus}
\citep{Li2026LoCoMoPlus} extends LoCoMo with a Cognitive subset
that probes the associative axis of \S\ref{sec:intro}: it injects
probes whose cue and answer item are bound only by narrative or
causal cues rather
than by lexical or semantic proximity, so that the cue item lies
far beyond any realistic top-$K$ under standard similarity
retrievers. Its Cognitive subset is the only new
contribution, and we report its score as the LoCoMo-Plus number
throughout \S\ref{sec:exp}.

\paragraph{Baselines.}
On LoCoMo (Table~\ref{tab:locomo}) we compare against Mem0
\citep{Chhikara2025Mem0}, Zep \citep{Rasmussen2025Zep},
Memobase\footnote{\url{https://www.memobase.io/}},
MemU\footnote{\url{https://github.com/NevaMind-AI/memU}},
Supermemory\footnote{\url{https://supermemory.ai/}}, MIRIX
\citep{Wang2025MIRIX}, MemOS \citep{Li2025MemOS} and HyperMem
\citep{Yue2026HyperMem}. On LoCoMo-Plus
(Table~\ref{tab:locomoplus}) we compare against the memory systems
Mem0, SeCom \citep{Pan2025SeCom}, A-Mem \citep{Xu2025AMEM}, MemOS
and HyperMem, plus two closed-source LLM reference baselines
(GPT-4o, Gemini-2.5-Pro), in line with \citet{Li2026LoCoMoPlus}.
Numbers taken from prior work vs.\ run by us are listed in the
captions of Table~\ref{tab:locomo} and Table~\ref{tab:locomoplus}.

\paragraph{Implementation Details.}
The dense encoder is
\texttt{bge-m3} \citep{Xiao2024BGE}. The three top-$K$ budgets
$(k^{\mathrm{T}}, k^{\mathrm{S}}, k^{\mathrm{I}})$ and the
trigger-union size follow the defaults swept in
\S\ref{sec:exp-hyperparam}. The memory-construction LLM is
GPT-4.1-mini in all our runs; we examine sensitivity to this
choice in Appendix~\ref{sec:appendix-repro}. For answer generation and
LLM-as-judge we follow each benchmark's official protocol: LoCoMo
uses GPT-4o-mini for both roles, with all LoCoMo numbers
averaged over three independent runs to align with the protocol
of \citet{Li2025MemOS}; LoCoMo-Plus uses GPT-4o for answer
generation and Gemini-2.5-Flash as judge under the binary
memory-awareness protocol of \citet{Li2026LoCoMoPlus}. Full hyperparameters in
Appendix~\ref{sec:appendix-repro}; prompt templates in
Appendix~\ref{sec:appendix-prompts}.

\subsection{Main Results on LoCoMo}
\label{sec:exp-locomo}

T-Mem reaches 80.26\% overall LLM-as-judge accuracy on
LoCoMo, 3.25 percentage points (pp) above the strongest
baseline HyperMem, and is the global maximum on five of the six
columns of the main block (Open Domain narrowly loses to MemOS
by 0.69~pp). Token-level F1 (51.96) corroborates this
ranking.

The four question-type columns line up with the design points
of \S\ref{sec:approach}. Single-hop and Temporal correspond to
the scene--item evidence layer, the Entity route of the
item-level Triggers, and the Scene Trigger. Multi-hop
corresponds to the Bridge route of the item-level Triggers and
the topic-label pre-filter, which together feed the candidate
pool with cross-scene evidence beyond the reach of a single
similarity hit. Open Domain corresponds to queries about a
speaker's standing traits, where T-Mem's Persona channel
complements the scene--item evidence.

% --- Table 4 (component ablations on LoCoMo and LoCoMo-Plus) -----
\begin{table}[t]
  \centering
  \small
  \setlength{\tabcolsep}{5pt}
  \begin{tabular}{l rr rr}
    \toprule
    \textbf{Configuration} & \multicolumn{2}{c}{\textbf{LoCoMo}} & \multicolumn{2}{c}{\textbf{LoCoMo-Plus}} \\
    \cmidrule(lr){2-3} \cmidrule(lr){4-5}
                & \textbf{\%} & $\boldsymbol{\Delta}$ & \textbf{\%} & $\boldsymbol{\Delta}$ \\
    \midrule
    \textbf{T-Mem} & \textbf{80.26} & -- & \textbf{74.81} & -- \\
    \midrule
    w/o SL      & 75.52 & $-4.74$ & 71.82 & $-2.99$ \\
    w/o IC      & 77.40 & $-2.86$ & 74.31 & $-0.50$ \\
    w/o ET + BT & 78.83 & $-1.43$ & 74.56 & $-0.25$ \\
    w/o PS      & 78.72 & $-1.54$ & 73.82 & $-0.99$ \\
    w/o TF      & 78.83 & $-1.43$ & 72.57 & $-2.24$ \\
    \midrule
    w/o ST      & 79.92 & $-0.34$ & 70.07 & $-4.74$ \\
    w/o HT      & 80.18 & $-0.08$ & 62.34 & $\boldsymbol{-12.47}$ \\
    w/o ST + HT & 79.86 & $-0.40$ & 52.62 & $\boldsymbol{-22.19}$ \\
    \bottomrule
  \end{tabular}
  \caption{Component ablations on LoCoMo and LoCoMo-Plus (Overall
    LLM-as-judge accuracy, \%). All eight ablations are run on the
    same memory system across both benchmarks.}
  \label{tab:ablation}
\end{table}

\subsection{Main Results on LoCoMo-Plus}
\label{sec:exp-locomoplus}

LoCoMo-Plus \citep{Li2026LoCoMoPlus} most directly tests the
Quadrant-II / III associative triggers.

T-Mem narrows the LoCoMo$\to$LoCoMo-Plus Gap to only
5.45~pp (Table~\ref{tab:locomoplus}) --- roughly five
times tighter than the strongest prior memory system HyperMem,
and nearly an order of magnitude tighter than the
Mem0 / SeCom / A-Mem cluster.

We attribute this contraction to the associative triggers,
with the scene-level Horizon
trigger carrying the bulk of the LoCoMo-Plus gain (ablation in
\S\ref{sec:exp-ablation}; the item-level Bridge trigger is hard
to register on this benchmark, see Limitations). As a
counterpoint, the closed-source LLMs in the top block of
Table~\ref{tab:locomoplus} still leave a Gap of roughly
45~pp under full-context input (zero-shot on the raw dialogue
history, no retrieval or memory mechanism), indicating that
backbone capacity alone cannot close the associative-axis
disconnect of LoCoMo-Plus.

% --- Figure 3 (radar chart of the eight ablation switches on both benchmarks, single column) ---
\begin{figure}[!t]
  \centering
  \includegraphics[width=0.8\linewidth]{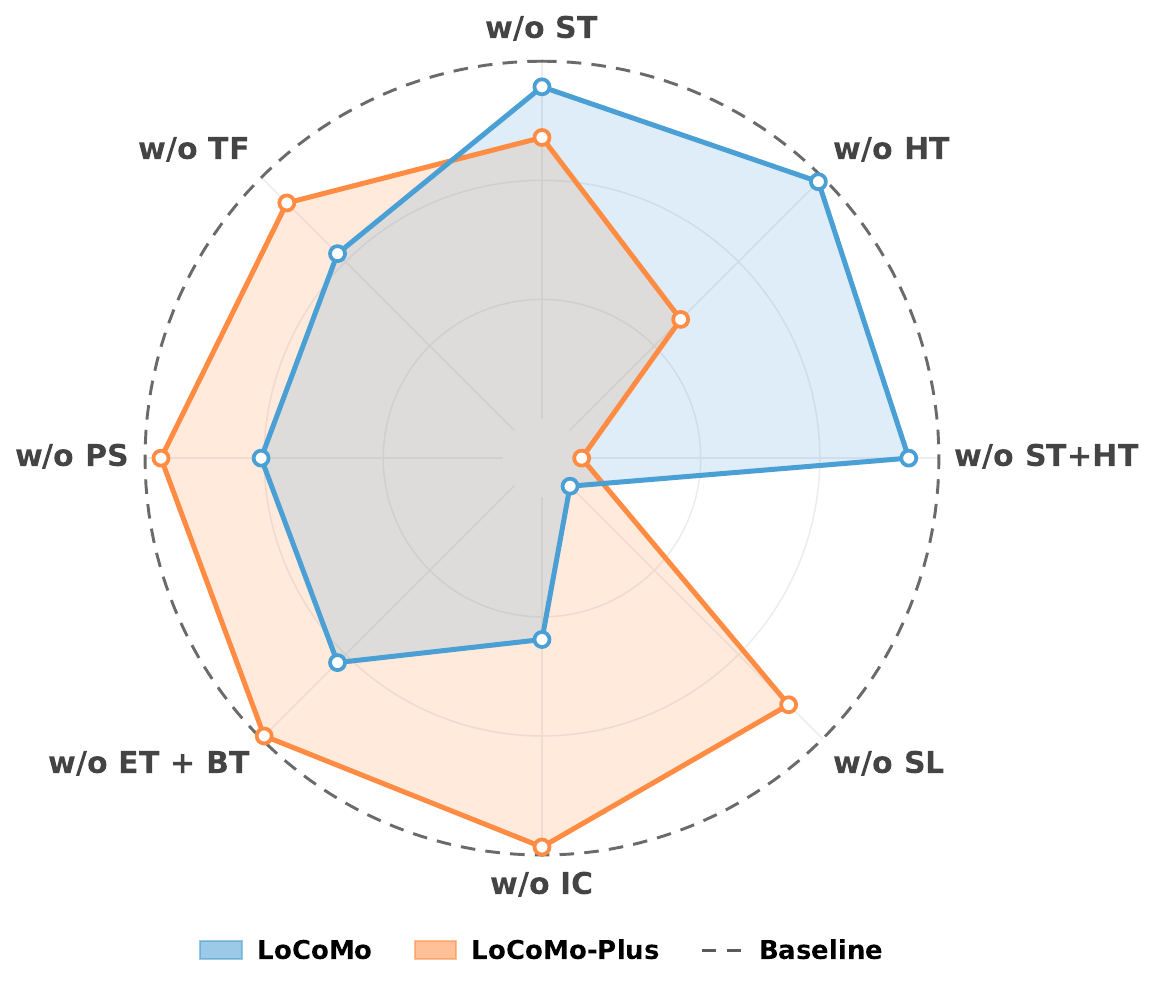}
  \caption{Component ablations on LoCoMo and LoCoMo-Plus.
    Each axis is one switch; radii are within-benchmark normalised
    so smaller radius means greater impact, with the most-affected
    switch on each dataset anchoring at the centre.}
  \label{fig:ablation}
\end{figure}

% --- Figure 4 ------------------------------------------------------
\begin{figure*}[t]
  \centering
  \includegraphics[width=\textwidth]{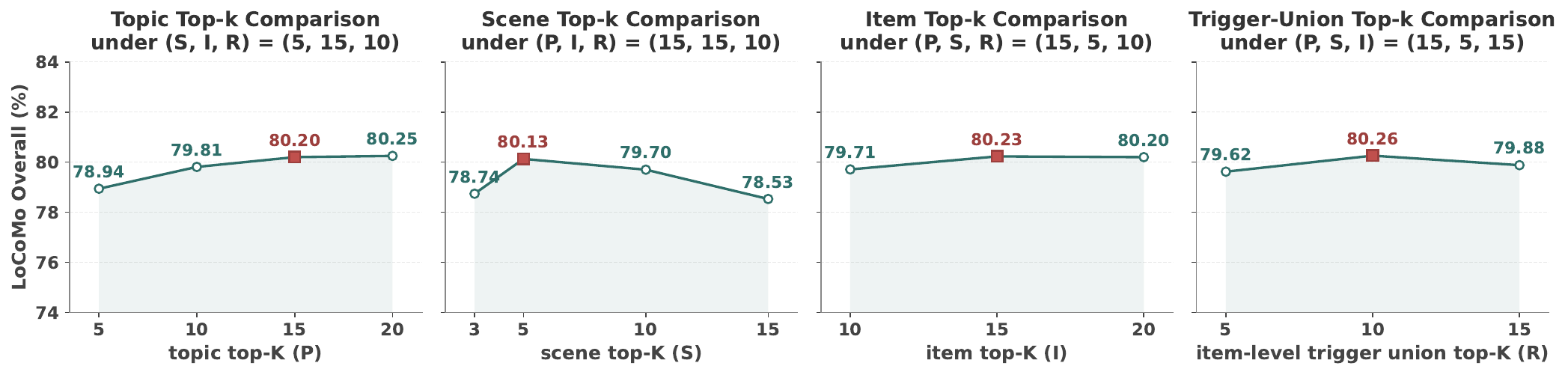}
  \caption{Hyperparameter sensitivity of T-Mem on LoCoMo Overall
    (\%). Each panel sweeps one of
    $\{P,S,I,R\}=\{k^{\mathrm{T}},k^{\mathrm{S}},k^{\mathrm{I}},R\}$
    (topic / scene / item / item-trigger union top-$K$) while
    holding the others at their defaults $(15,5,15,10)$;
    the default operating point is marked with a red square.}
  \label{fig:hyperparam}
\end{figure*}

\subsection{Ablation Study}
\label{sec:exp-ablation}

Table~\ref{tab:ablation} and Figure~\ref{fig:ablation} report
Overall LLM-as-judge accuracy across eight ablations on both
benchmarks; the eight switches partition T-Mem's components
into a descriptive-axis group (scene / item evidence layer,
Entity\,+\,Bridge triggers, Persona, topic prefilter) and an
associative-axis group (the two scene-level triggers Scene and
Horizon), and the asymmetry between the two columns carries the
argument. The eight switches, in table order, are SL (Scene
Layer), IC (Item Channel), ET+BT (Entity + Bridge Triggers),
PS (Persona), TF (Topic-label Filter), and the two scene-level
triggers ST (Scene Trigger) and HT (Horizon Trigger). All
``w/o XX'' labels below correspond directly to rows of
Table~\ref{tab:ablation}.

The top block (SL / IC / ET+BT / PS / TF) shows a strong
contrast between the two benchmarks: drops of $1.4$--$4.7$~pp
on LoCoMo, but only $0.25$--$2.99$~pp on LoCoMo-Plus. This
asymmetry is rooted in the LoCoMo-Plus format itself, which is
dominated by scene-level associative evidence; item-level
components, including the Bridge trigger, are hard to
register on this benchmark even when retrieved and supplied to
the QA prompt.

The bottom block (ST / HT, individually and jointly) inverts the
picture: all three configurations move Overall by under half
a point on LoCoMo, but collapse it by up to $-22.19$~pp on
LoCoMo-Plus, with ST\,+\,HT exceeding every descriptive-axis
switch on that subset by more than $7\times$. These
scene-level triggers therefore sit at the bottom of
LoCoMo's effect ranking and at the top of LoCoMo-Plus's,
consistent with their role in \S\ref{sec:approach}. This
asymmetry, silent on LoCoMo and decisive on LoCoMo-Plus, is the
empirical form of \S\ref{sec:intro}'s diagnosis that a system
optimised for LoCoMo is implicitly optimised for staying inside
the similarity neighbourhood. An axis benchmarked along
similarity alone cannot, by construction, register the cost of
removing the associative axis.

\subsection{Hyperparameter Analysis}
\label{sec:exp-hyperparam}

We sweep the four budgets defined in
\S\ref{sec:approach-retrieval} (Figure~\ref{fig:hyperparam}):
topic top-$K$ ($k^{\mathrm{T}}$),
scene top-$K$ ($k^{\mathrm{S}}$), item top-$K$ ($k^{\mathrm{I}}$),
and the item-trigger union top-$K$ ($R$). We vary one at a time
with others fixed; the QA configuration
matches Table~\ref{tab:locomo}'s main rows.

Two of the four sweeps are monotone-saturating, two are
sweet-spot-shaped: topic top-$K$ saturates by 15 (with no
measurable further gain at 20) and item top-$K$ at 15, while scene
top-$K$ and the item-trigger union top-$K$ peak at 5 and 10
respectively. The default operating point
$(k^{\mathrm{T}},k^{\mathrm{S}},k^{\mathrm{I}},R)=(15,5,15,10)$
sits at the saturation knee or sweet spot of each sweep,
indicating that T-Mem is not finely tuned to a narrow operating
regime.

% --- Figure 5 ------------------------------------------------------
\begin{figure}[!t]
  \centering
  \includegraphics[width=\columnwidth]{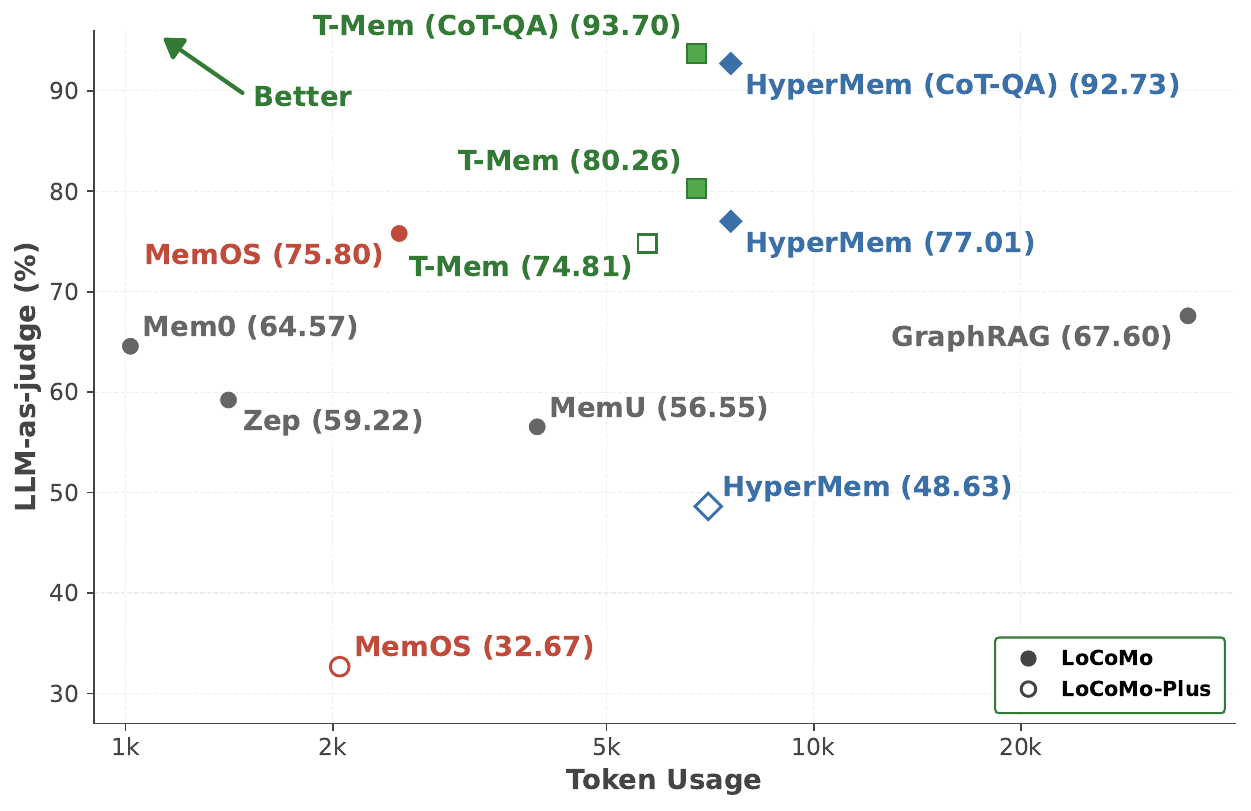}
  \caption{Token usage versus accuracy on LoCoMo and LoCoMo-Plus.
    The $x$-axis is average input tokens per query (log scale).
    Token-usage numbers for systems other than T-Mem, HyperMem,
    and MemOS on LoCoMo-Plus are taken from the HyperMem paper
    \citep{Yue2026HyperMem}.}
  \label{fig:efficiency}
\end{figure}

\subsection{Efficiency Analysis}
\label{sec:exp-cost}

\paragraph{Construction Cost.}
Memory construction is a one-time offline cost, separate from
the answer-time budget analysed below. Across all 10 LoCoMo
conversations, T-Mem's construction pipeline (scene
segmentation, topic assignment, item extraction, and the four
trigger families) issues 9,689 GPT-4.1-mini calls and consumes
15.60M tokens, averaging $\approx$969 calls and $\approx$1.56M
tokens per conversation. Table~\ref{tab:construction-cost}
places this alongside recent memory systems. T-Mem's total
(15.60M) exceeds Mem0's (12.20M); we attribute the gap to the
two associative trigger families (Bridge, Horizon), whose
benefit is validated on LoCoMo-Plus (\S\ref{sec:exp-ablation})
but limited on LoCoMo, where factual recall dominates.
We view this overhead not as inefficiency but as the price of
extending the trigger space along the associative axis
(\S\ref{sec:approach-construction-triggers}): it is precisely
what equips T-Mem with associative recall, a capability none of
the purely fact-based systems in
Table~\ref{tab:construction-cost} possesses.

\begin{table}[t]
  \centering
  \small
  \setlength{\tabcolsep}{5pt}
  \begin{tabular}{l rrrr}
    \toprule
    \textbf{Method} & \textbf{In(M)} & \textbf{Out(M)} & \textbf{Sum(M)} & \textbf{Calls} \\
    \midrule
    LightRAG              & 10.01 & 1.92 & 11.93 & 13,576 \\
    A-Mem                 & 9.13  & 2.37 & 11.49 & 11,754 \\
    Mem0                  & 10.96 & 1.24 & 12.20 & 9,181  \\
    MemoryOS              & 1.89  & 0.94 & 2.87  & 5,534  \\
    Mem0g                 & 33.51 & 2.31 & 35.83 & 53,514 \\
    StructMem             & 1.50  & 0.44 & 1.94  & 1,056  \\
    \textbf{T-Mem (Ours)} & \textbf{13.58} & \textbf{2.02} & \textbf{15.60} & \textbf{9,689} \\
    \bottomrule
  \end{tabular}
  \caption{Memory-construction cost on LoCoMo (input / output /
    total tokens in millions; LLM calls), aggregated over all 10
    conversations. Numbers for LightRAG, A-Mem, Mem0, MemoryOS,
    Mem0g and StructMem are taken from \citet{Xu2026StructMem}.}
  \label{tab:construction-cost}
\end{table}

\paragraph{Answer-Time Efficiency.}
A retrieval-augmented memory architecture must justify its added
cost. Figure~\ref{fig:efficiency} contrasts each system's input-token
usage against its overall LLM-as-judge accuracy on both
benchmarks (log-scaled $x$-axis). T-Mem stands out on both: at a
token budget below HyperMem's, it simultaneously reaches higher
accuracy on LoCoMo and on LoCoMo-Plus. The lower-token cluster
(Mem0, Zep, MemOS) trades token cost for a clear accuracy
deficit on LoCoMo and collapses on LoCoMo-Plus. Under the
CoT-QA configuration, T-Mem likewise leads HyperMem at the same
token budget, so the lead persists across both QA pipelines.

\section{Conclusion}
\label{sec:conclusion}

Current long-term conversational memory systems cover only one
mode of recall, the one driven by similarity between a query and
stored content; the other mode, where query and memory are bound
by latent semantic association, is structurally left out. We
propose \textbf{T-Mem}, which places one trigger family
per quadrant of the $2\!\times\!2$ recall design space, so that
every memory remains reachable from both descriptively similar
and associatively relevant queries. The empirical payoff is
visible across both LoCoMo and LoCoMo-Plus, where T-Mem reaches
state-of-the-art and collapses the cross-benchmark gap into a
single-digit margin. This is the engineering form of a
long-standing claim from cognitive science:
\emph{a long-term memory system earns its adaptive value not by
archiving the dialogue stream faithfully, but by anticipating, at
write time, the future cues under which its contents will need
to be reached.}

% =====================================================================
% Limitations
% =====================================================================
\section*{Limitations}

The writing pipeline (scene segmentation, item
extraction, four-family trigger instantiation, and Persona
summarisation) relies on a memory-construction LLM strong enough
to follow structured-output instructions reliably. Memory is
also built offline, leaving incremental update and consolidation,
as well as reinforcement-learning memory management
\citep{Yan2025MemoryR1,Wang2025MemAlpha}, to future work.
Finally, LoCoMo-Plus's Cognitive subset is to our knowledge the
only public benchmark that directly probes the associative axis
we target; its cue is itself a short dialogue, which
structurally favours scene-granularity recall and gives the
item-level Bridge trigger no lever. A fact-level cognitive
benchmark with a single-fact cue would round out the other half
of the evidence.

% =====================================================================
% Ethics Statement
% =====================================================================
\section*{Ethics Statement}

All experiments in this paper use LoCoMo \citep{Maharana2024LoCoMo}
and LoCoMo-Plus \citep{Li2026LoCoMoPlus}, two publicly released
benchmarks whose dialogues are synthetically generated and
contain no personal, sensitive, or identifiable information about
real individuals; we process no real-user data and did not
collect any new human-subject data. We nonetheless recognise
that a production deployment of T-Mem, unlike our benchmark-only
setting, would store richly structured, trigger-annotated
profiles of real users' conversations over time. Such a
deployment would warrant standard privacy-by-design safeguards
(e.g., access control, data minimisation, and user rights to
review or delete stored memory) that are outside the scope of
the architecture study presented here.

% =====================================================================
% Acknowledgments
% =====================================================================
\section*{Acknowledgments}

We thank the anonymous reviewers and area chair for their
constructive comments. We also thank our colleagues at Tencent
for helpful discussions during the course of this work.

% Bibliography entries for the entire Anthology, followed by custom entries
%\bibliography{custom,anthology-overleaf-1,anthology-overleaf-2}

% Custom bibliography entries only
\bibliography{custom}

\appendix

% =====================================================================
% Appendix A.1 -- Algorithms
% =====================================================================
\section{Algorithms}
\label{sec:appendix-algorithm}

Algorithms~\ref{alg:retrieval-full}--\ref{alg:indexing} together
expand the entire T-Mem pipeline into stage-by-stage pseudocode:
the online retrieval cascade
(\S\ref{sec:approach-retrieval}, including the trigger-recall
calls invoked therein), offline construction
(\S\ref{sec:approach-construction}), and offline indexing
(\S\ref{sec:approach-indexing}). Algorithm~\ref{alg:retrieval-full}
in particular spells out the internals of the
$\mathrm{SceneCue}(\cdot)$ and $\mathrm{TrigRecall}(\cdot)$ calls
abstracted in Algorithm~\ref{alg:retrieval-main} of the main paper.
Algorithm~\ref{alg:topic-assignment} expands Stage~2 of the
construction pipeline (the $\mathrm{TopicLLM}$ call in
Algorithm~\ref{alg:construction}) into its three subroutines
--- batched matching, new-topic creation, and topic update.

% --- Algorithm A.1 : online retrieval (full) -----------------------
\begin{algorithm}[!h]
\small
\caption{T-Mem online retrieval (full): the
  $\mathrm{SceneCue}$ and $\mathrm{TrigRecall}$ calls of
  Algorithm~\ref{alg:retrieval-main} are unfolded inline.}
\label{alg:retrieval-full}
\begin{algorithmic}[1]
\Require query $q$;\, budgets $k^{\mathrm{T}}\!,k^{\mathrm{S}}\!,k^{\mathrm{I}}$;\, item-trigger gate $\tau{=}0.85$
\Ensure  retrieval context handed to the QA LLM
\Statex \textbf{Stage 1: topic-label filtering}
\State $\mathcal{R}_{T}\!\gets\!\mathrm{top}_{k^{\mathrm{T}}}\!\bigl(\mathrm{RRF}(q,\mathcal{V}^{T})\bigr)$
       \Comment{Eq.~\ref{eq:rrf} over BM25, dense}
\State $\mathcal{P}_{S}\!\gets\!\{v^{S}\!:\!(v^{T},v^{S})\!\in\!\mathcal{E}^{TS},\, v^{T}\!\in\!\mathcal{R}_{T}\}$
\Statex \textbf{Stage 2: scene selection (similarity $\cup$ associative)}
\State \Comment{$\mathrm{SceneCue}(q)$ unfolded:}
\State $r^{\mathrm{dlg}},r^{\mathrm{scn}},r^{\mathrm{hor}}\gets\mathrm{Cos}(q,\mathbf{e}^{\mathrm{dlg}}),\mathrm{Cos}(q,\mathbf{e}^{\mathrm{scn}}),\max_{s}\mathrm{Cos}(q,\mathbf{e}^{\mathrm{hor}}_{s})$
\State $\mathcal{A}_{S}\!\gets\!\mathrm{top}\!\bigl(\mathrm{RRF}(r^{\mathrm{dlg}},r^{\mathrm{scn}},r^{\mathrm{hor}})\bigr)$
       \Comment{Eq.~\ref{eq:rrf} with $M{=}3$ $\to$ host scenes}
\State $\mathcal{C}_{2}\!\gets\!\mathcal{P}_{S}\cup\mathcal{A}_{S}$
\State $\mathcal{R}_{S}\!\gets\!\mathrm{top}_{k^{\mathrm{S}}}\!\bigl(\mathrm{RRF}(q,\mathcal{C}_{2})\bigr)$
\Statex \textbf{Stage 3: item selection (scene-bound $\cup$ trigger-reached)}
\State $\mathcal{P}_{I}\!\gets\!\{v^{I}\!:\!(v^{S},v^{I})\!\in\!\mathcal{E}^{SI},\, v^{S}\!\in\!\mathcal{R}_{S}\}$
\State \Comment{$\mathrm{TrigRecall}(q;\tau)$ unfolded:}
\For{$v^{I}\in\mathcal{V}^{I}$}
  \State $s_{v^{I}}\!\gets\!\mathrm{nanmax}\bigl(\mathrm{Cos}(q,\mathbf{e}^{\mathrm{con}}_{v^{I}}),$
         \Statex \hspace{3.2em}$\mathrm{Cos}(q,\mathbf{e}^{\mathrm{brg}}_{v^{I}}),\mathrm{Cos}(q,\mathbf{e}^{\mathrm{joint}}_{v^{I}})\bigr)$
\EndFor
\State $\mathcal{A}_{I}\!\gets\!\{v^{I}\!:\!s_{v^{I}}\!\geq\!\tau\}$
       \Comment{hard cosine gate $\to$ host items}
\State $\mathcal{C}_{3}\!\gets\!\mathcal{P}_{I}\cup\mathcal{A}_{I}$
\State $\mathcal{R}_{I}\!\gets\!\mathrm{top}_{k^{\mathrm{I}}}\!\bigl(\mathrm{RRF}(q,\mathcal{C}_{3})\bigr)$
\Statex \textbf{Persona $+$ answer generation}
\State $X\!\gets\!\mathrm{Persona}(\mathrm{speaker}(q))$
\State \Return $\mathrm{QA\_LLM}\!\bigl(q,\,\mathcal{R}_{S},\,\mathcal{R}_{I},\,X\bigr)$
\end{algorithmic}
\end{algorithm}

% --- Algorithm A.2 : offline construction --------------------------
\begin{algorithm}[!h]
\small
\caption{T-Mem offline construction.}
\label{alg:construction}
\begin{algorithmic}[1]
\Require dialogue stream $X=(x_{t})_{t=1}^{T}$
\Ensure  memory $\mathcal{M}=(\mathcal{V}^{T}\!\cup\!\mathcal{V}^{S}\!\cup\!\mathcal{V}^{I},\,\mathcal{E}^{TS}\!\cup\!\mathcal{E}^{SI},\,\mathcal{T}^{*},\,\mathcal{X})$
\State $\mathcal{V}^{S},\mathcal{V}^{T},\mathcal{V}^{I}\gets\varnothing$;\;
       $\mathcal{E}^{TS},\mathcal{E}^{SI}\gets\varnothing$;\;
       $H\gets[\,]$
\Statex \textbf{Stage 1: scene segmentation}
\For{$x_{t}\in X$}
  \State append $x_{t}$ to buffer $H$
  \If{$\mathrm{BoundaryLLM}(H)=\textsc{close}$}
    \State emit $v^{S}\!=\!\mathrm{ScenePack}(H)$;\;
           $\mathcal{V}^{S}\!\gets\!\mathcal{V}^{S}\cup\{v^{S}\}$
    \State $H\gets[\,]$
  \EndIf
\EndFor
\Statex \textbf{Stage 2: topic assignment}
\For{$v^{S}_{\text{new}}\in\mathcal{V}^{S}$ (in arrival order)}
  \State $\mathcal{A}\gets\mathrm{TopicLLM}(v^{S}_{\text{new}},\,\mathcal{V}^{T})$
         \Comment{batched per-label admission}
  \If{$\mathcal{A}=\varnothing$}
    \State open new $v^{T}$ from $v^{S}_{\text{new}}$;\;
           $\mathcal{V}^{T}\!\gets\!\mathcal{V}^{T}\cup\{v^{T}\}$;\;
           add $(v^{T},v^{S}_{\text{new}})$ to $\mathcal{E}^{TS}$
  \Else
    \For{$v^{T}\in\mathcal{A}$}
      \State add $(v^{T},v^{S}_{\text{new}})$ to $\mathcal{E}^{TS}$
      \State refresh metadata of $v^{T}$ over its scenes
    \EndFor
  \EndIf
\EndFor
\Statex \textbf{Stage 3: item extraction}
\For{$v^{T}\in\mathcal{V}^{T}$}
  \State $\mathcal{V}^{S}_{t}\gets\{v^{S}\!:(v^{T},v^{S})\!\in\!\mathcal{E}^{TS}\}$
  \State $(\mathcal{I}^{\mathrm{atom}},\mathcal{I}^{\mathrm{conn}})\gets\mathrm{ItemLLM}(v^{T},\mathcal{V}^{S}_{t})$
  \For{$v^{I}\in\mathcal{I}^{\mathrm{atom}}$}
    \State $\mathcal{V}^{I}\!\gets\!\mathcal{V}^{I}\cup\{v^{I}\}$;\;
           add single $(v^{S}_{\mathrm{src}}(v^{I}),v^{I})$ to $\mathcal{E}^{SI}$
  \EndFor
  \For{$v^{I}\in\mathcal{I}^{\mathrm{conn}}$}
    \State $\mathcal{V}^{I}\!\gets\!\mathcal{V}^{I}\cup\{v^{I}\}$;\;
           add edges $\{(v^{S},v^{I})\!:v^{S}\!\in\!\mathcal{V}^{S}_{\mathrm{src}}(v^{I})\}$ to $\mathcal{E}^{SI}$
  \EndFor
\EndFor
\Statex \textbf{Stage 4: trigger instantiation}
\For{$v^{I}\in\mathcal{V}^{I}$}
  \State $(\mathcal{T}^{\mathrm{Ent}}_{v^{I}},\mathcal{T}^{\mathrm{Brg}}_{v^{I}},r_{v^{I}})\gets\mathrm{ItemTrigLLM}(v^{I})$
         \Comment{Q\,I, Q\,II $+$ rationale}
\EndFor
\For{$v^{S}\in\mathcal{V}^{S}$}
  \State $(\mathcal{T}^{\mathrm{Scn}}_{v^{S}},\mathcal{T}^{\mathrm{Hor}}_{v^{S}})\gets\mathrm{SceneTrigLLM}(v^{S})$
         \Comment{Q\,IV, Q\,III}
\EndFor
\Statex \textbf{Persona (per speaker $u$, in parallel)}
\State $\mathcal{X}\gets\{\,(u,\mathrm{PersonaLLM}(\{x_{t}\!:\!\mathrm{spk}(x_{t})\!=\!u\}))\,\}_{u}$
\State \Return $\mathcal{M}$
\end{algorithmic}
\end{algorithm}

% --- Algorithm A.3 : topic assignment (subroutines) ----------------
\begin{algorithm}[!h]
\small
\caption{T-Mem topic assignment: the $\mathrm{TopicLLM}$ call of
  Algorithm~\ref{alg:construction} (Stage~2) unfolded into three
  subroutines.}
\label{alg:topic-assignment}
\begin{algorithmic}[1]
\Require new scene $v^{S}$;\, existing topic pool $\mathcal{V}^{T}$;\, batch size $b$
\Ensure  updated topic pool $\mathcal{V}^{T}$ and edge set $\mathcal{E}^{TS}$
\If{$\mathcal{V}^{T}=\varnothing$}
  \State $v^{T}\!\gets\!\textsc{CreateNewTopic}(\{v^{S}\})$
  \State $\mathcal{V}^{T}\!\gets\!\mathcal{V}^{T}\cup\{v^{T}\}$;\;
         add $(v^{T},v^{S})$ to $\mathcal{E}^{TS}$
  \State \Return
\EndIf
\Statex \textbf{Match phase (batched).}
\State $\mathcal{A}\gets\varnothing$
\For{each batch $\mathcal{B}\subseteq\mathcal{V}^{T}$ of size $\le b$}
  \State $\mathcal{A}\gets\mathcal{A}\cup\textsc{MatchBatch}(v^{S},\mathcal{B})$
         \Comment{LLM admits $v^{S}$ to multiple topics}
\EndFor
\If{$\mathcal{A}=\varnothing$}
  \State $v^{T}\!\gets\!\textsc{CreateNewTopic}(\{v^{S}\})$
  \State $\mathcal{V}^{T}\!\gets\!\mathcal{V}^{T}\cup\{v^{T}\}$;\;
         add $(v^{T},v^{S})$ to $\mathcal{E}^{TS}$
  \State \Return
\EndIf
\Statex \textbf{Update phase.}
\For{each $v^{T}\in\mathcal{A}$}
  \State $v^{T}\!\gets\!\textsc{UpdateTopic}(v^{T},v^{S})$
  \State add $(v^{T},v^{S})$ to $\mathcal{E}^{TS}$
\EndFor
\Statex \rule{\linewidth}{0.4pt}
\Statex \textbf{Subroutine} $\textsc{MatchBatch}(v^{S},\mathcal{B})$:
\Statex \quad ask the matcher LLM, for each $v^{T}\!\in\!\mathcal{B}$, whether $v^{S}$ continues the same specific event thread as $v^{T}$;
\Statex \quad return the subset for which the LLM answers \textsc{true}.
\Statex \textbf{Subroutine} $\textsc{CreateNewTopic}(\{v^{S}\})$:
\Statex \quad ask an extractor LLM for a specific title and a keyword list from $v^{S}$;
\Statex \quad store $\mathrm{summary}\!=\!\mathrm{title}{\times}2\,\Vert\,\mathrm{keywords}$ for the BM25 index of \S\ref{sec:approach-indexing}.
\Statex \textbf{Subroutine} $\textsc{UpdateTopic}(v^{T},v^{S})$:
\Statex \quad ask an updater LLM to fold $v^{S}$ into $v^{T}$ while keeping the title's specific identity stable; merge new keywords into the existing list and recompute $\mathrm{summary}$.
\end{algorithmic}
\end{algorithm}

% --- Algorithm A.4 : offline indexing ------------------------------
\begin{algorithm}[!h]
\small
\caption{T-Mem offline indexing.}
\label{alg:indexing}
\begin{algorithmic}[1]
\Require memory $\mathcal{M}$ with node sets $\mathcal{V}^{T},\mathcal{V}^{S},\mathcal{V}^{I}$ and trigger sets $\mathcal{T}^{\mathrm{Ent}},\mathcal{T}^{\mathrm{Brg}},\mathcal{T}^{\mathrm{Scn}},\mathcal{T}^{\mathrm{Hor}}$;
        encoder $\mathrm{Enc}(\cdot)$
\Ensure  BM25 corpus $\mathcal{B}$, dense tables $\mathcal{D}^{\bullet}$, multi-view trigger indices
\Statex \textbf{Node-level lexical and dense}
\For{$\bullet\in\{T,S,I\}$ and $v\in\mathcal{V}^{\bullet}$}
  \State register $\mathrm{Concat}_{\mathrm{w}}(v)$ into shared BM25 corpus $\mathcal{B}$
         \Comment{per-type weighted field concat}
  \State $\mathcal{D}^{\bullet}[v]\gets\mathrm{Enc}(v)$
\EndFor
\Statex \textbf{Item-level multi-view trigger index}
\For{$v^{I}\in\mathcal{V}^{I}$ with $(\mathcal{T}^{\mathrm{Ent}}_{v^{I}},\mathcal{T}^{\mathrm{Brg}}_{v^{I}},r_{v^{I}})$}
  \State $\mathbf{e}^{\mathrm{con}}_{v^{I}}\!\gets\!\mathrm{Enc}(\mathcal{T}^{\mathrm{Ent}}_{v^{I}})$;\;
         $\mathbf{e}^{\mathrm{brg}}_{v^{I}}\!\gets\!\mathrm{Enc}(\mathcal{T}^{\mathrm{Brg}}_{v^{I}})$;\;
         $\mathbf{e}^{\mathrm{joint}}_{v^{I}}\!\gets\!\mathrm{Enc}(\mathcal{T}^{\mathrm{Ent}}_{v^{I}}\!\Vert\mathcal{T}^{\mathrm{Brg}}_{v^{I}}\!\Vert r_{v^{I}})$
  \State store $(v^{I},\mathbf{e}^{\mathrm{con}},\mathbf{e}^{\mathrm{brg}},\mathbf{e}^{\mathrm{joint}})$;\;
         empty fields written as $\mathrm{NaN}$ sentinel
\EndFor
\Statex \textbf{Scene-level multi-view trigger index}
\For{$v^{S}\in\mathcal{V}^{S}$ with $(\mathcal{T}^{\mathrm{Scn}}_{v^{S}},\mathcal{T}^{\mathrm{Hor}}_{v^{S}})$}
  \State $\mathbf{e}^{\mathrm{dlg}}_{v^{S}}\!\gets\!\mathrm{Enc}(\mathrm{dialogue}(v^{S}))$;\;
         $\mathbf{e}^{\mathrm{scn}}_{v^{S}}\!\gets\!\mathrm{Enc}(\mathcal{T}^{\mathrm{Scn}}_{v^{S}})$;\;
         $\mathbf{e}^{\mathrm{hor}}_{v^{S}}\!\gets\!\{\mathrm{Enc}(s)\!:\!s\!\in\!\mathcal{T}^{\mathrm{Hor}}_{v^{S}}\}$
\EndFor
\State \Return $\mathcal{B}$, $\{\mathcal{D}^{\bullet}\}$, item-trigger index, scene-trigger index
\end{algorithmic}
\end{algorithm}

% =====================================================================
% Appendix B -- Prompts
% =====================================================================
\section{Prompts}
\label{sec:appendix-prompts}

We summarise the prompt templates that govern T-Mem's pipeline
and evaluation; the full prompt templates, including all few-shot
exemplars, are released with the codebase. The four T-Mem
templates are reproduced in
Figures~\ref{fig:prompt-item}--\ref{fig:prompt-qa} (collected at
the end of the appendix): item extraction
(Figure~\ref{fig:prompt-item}), item-level Triggers covering the
Entity and Bridge routes
(Figure~\ref{fig:prompt-itemtrigger}), scene-level Triggers
covering the Scene and Horizon layers
(Figure~\ref{fig:prompt-scenetrigger}), and the matched LoCoMo QA
prompt (Figure~\ref{fig:prompt-qa}). For the QA-pipeline
decomposition of \S\ref{sec:appendix-repro}, we additionally
reproduce HyperMem's 7-step CoT QA prompt
(Figure~\ref{fig:prompt-qa-hypermem}), so that the two QA prompts
underlying the matched and the $\dagger$ rows of
Table~\ref{tab:locomo} can be read side by side.

% =====================================================================
% Appendix C -- Case studies (one per quadrant)
% =====================================================================
\section{Case studies}
\label{sec:appendix-perquad}

We illustrate T-Mem's behaviour with five representative cases
drawn from real test instances: four from LoCoMo (covering the
single-hop, multi-hop, temporal, and open domain question types
of \citealp{Maharana2024LoCoMo}) and one from LoCoMo-Plus
(covering the cognitive cue-continuation regime of
\citealp{Li2026LoCoMoPlus}). Each case shows the recalled
evidence, the query, the gold answer, T-Mem's prediction, and
the mechanism behind the correct call. The five cases are
collected at the end of the appendix as
Figures~\ref{fig:case-singlehop}--\ref{fig:case-cognitive}:
single-hop (Figure~\ref{fig:case-singlehop}), temporal
(Figure~\ref{fig:case-temporal}), multi-hop
(Figure~\ref{fig:case-multihop}), open domain
(Figure~\ref{fig:case-opendomain}), and cognitive
(Figure~\ref{fig:case-cognitive}).

% =====================================================================
% Appendix D -- Terminology: trigger query (LoCoMo-Plus) vs. trigger (T-Mem)
% =====================================================================
\section{Terminology: ``trigger query'' vs.\ trigger}
\label{sec:appendix-terminology}

Our \emph{triggers} (Entity, Bridge, Scene, Horizon;
\S\ref{sec:approach}) are memory-side indexing objects attached at
write time. The phrase ``trigger query'' in
\citet{Li2026LoCoMoPlus} refers, by contrast, to a query-side probe
issued after a time gap. The two usages target opposite ends of
the same regime and are not in conflict.

% =====================================================================
% Appendix E -- Reproducibility
% =====================================================================
\section{Reproducibility}
\label{sec:appendix-repro}

\paragraph{Decomposing the two QA pipelines.}
The 15.72-pp accuracy gap between HyperMem's reported 92.73 and our
re-run of HyperMem under the official LoCoMo pipeline (77.01) is
driven jointly by two confounded factors bundled inside HyperMem's
own QA pipeline. The first is a stronger answer-generation LLM
(GPT-4.1-mini vs.\ GPT-4o-mini); the second is a 7-step CoT QA
prompt that encourages long, exhaustive answers, against the
official LoCoMo prompt that caps answers at 5--6 words. The two
prompts can be read side by side in
Figures~\ref{fig:prompt-qa} and~\ref{fig:prompt-qa-hypermem}; the
full HyperMem template is available in their released code. To
disentangle the two factors we
freeze HyperMem's retrieval output on all 1{,}540 LoCoMo questions
and vary only the QA stage along these two axes. This defines
three configurations: (A)~HyperMem's reported setup (GPT-4.1-mini +
7-step CoT); (B)~the official LoCoMo setup (GPT-4o-mini +
official prompt); and (C)~a controlled rerun that swaps only the
answer-generation LLM, keeping the 7-step CoT prompt
(GPT-4o-mini + 7-step CoT). Mean prediction length is 76.95 tokens
under (A), 45.71 under (C), and 5.87 under (B), so prompt and
model each move length by a comparable factor and stack
multiplicatively. Crucially, both factors push answers longer in
exactly the way that benefits an LLM-as-judge over a token-level
F1 metric, which is why the $\dagger$ rows in
Table~\ref{tab:locomo} simultaneously show a $\sim$16-pp accuracy
inflation and an $\sim$30-pp F1 collapse against the matched-pipeline
rows.

\paragraph{Worked case --- temporal disambiguation.}
The mechanism by which the 7-step CoT prompt converts retrieval
output into LLM-judge gains is most cleanly visible on temporal
questions whose evidence supports several plausible answers.
Figure~\ref{fig:case-tokyo} contrasts the two pipelines on the same
(retrieval, question) pair.

\paragraph{Takeaway.}
The 7-step CoT prompt is not directly comparable with prior
LoCoMo numbers tied to the 5--6-word official prompt. We therefore
report T-Mem under both: 80.26 in the matched-pipeline rows of
Table~\ref{tab:locomo}, and 93.70 in the $\dagger$ rows. T-Mem
leads under both.

\paragraph{Build-Model Sensitivity.}
Our main results (\S\ref{sec:exp-benchmarks}) fix the
memory-construction LLM at GPT-4.1-mini. To test how much of
T-Mem's gain is attributable to this specific choice rather
than to the memory architecture itself, we swap in the
open-weight Qwen3-32B while keeping QA and judge fixed at
GPT-4o-mini (Table~\ref{tab:buildmodel-qwen}; 3 judge runs,
1,540 questions): Overall drops by 4.81 pp (80.26 $\to$ 75.45),
with the largest degradation on Multi-hop ($-8.39$ pp),
consistent with multi-hop questions placing the heaviest
cross-scene reasoning demand on the construction LLM. The
Qwen3-32B build still substantially exceeds mainstream memory
baselines such as Mem0 and Memobase in Table~\ref{tab:locomo}.

\begin{table}[t]
  \centering
  \small
  \setlength{\tabcolsep}{3pt}
  \begin{tabular}{l ccccc}
    \toprule
    \textbf{Build Model} & \textbf{Overall} & \textbf{Single} & \textbf{Multi} & \textbf{Temp} & \textbf{Open} \\
    \midrule
    GPT-4.1-mini            & \textbf{80.26} & \textbf{85.97} & \textbf{69.15} & \textbf{82.55} & \textbf{55.21} \\
    Qwen3-32B               & 75.45 & 83.31 & 60.76 & 75.08 & 51.04 \\
    \midrule
    $\Delta$                & $-4.81$ & $-2.66$ & $-8.39$ & $-7.47$ & $-4.17$ \\
    \bottomrule
  \end{tabular}
  \caption{Build-model swap on LoCoMo (QA / judge fixed at
    GPT-4o-mini). Overall LLM-as-judge accuracy, \%.}
  \label{tab:buildmodel-qwen}
\end{table}

Table~\ref{tab:buildmodel-cross} further crosses the build and
QA models (judge fixed at GPT-4o-mini). Fixing the QA model and
upgrading the build model from GPT-4.1 to GPT-5.1 yields only
$+3.05$~pp (78.31 $\to$ 81.36); fixing the build model and
upgrading the QA model instead yields a comparable
$+4.20$~pp (GPT-4.1 build) or $+3.49$~pp (GPT-5.1 build).
Relative rankings across configurations stay stable, indicating
that stronger models raise T-Mem's absolute accuracy without
altering the relative contribution attributable to the memory
architecture itself.

\begin{table}[t]
  \centering
  \small
  \setlength{\tabcolsep}{4pt}
  \begin{tabular}{l ccccc}
    \toprule
    \textbf{Build / QA} & \textbf{Overall} & \textbf{Single} & \textbf{Multi} & \textbf{Temp.} & \textbf{Open} \\
    \midrule
    4.1 / 4o-mini & 78.31 & 83.67 & 68.44 & 82.66 & 45.83 \\
    4.1 / 4.1     & 82.51 & 88.19 & 73.52 & 85.88 & 47.92 \\
    5.1 / 4o-mini & 81.36 & 87.75 & 67.97 & 84.84 & 53.12 \\
    5.1 / 5.1     & 84.85 & 90.96 & 74.70 & 85.98 & 57.29 \\
    \bottomrule
  \end{tabular}
  \caption{Build $\times$ QA model cross-comparison on LoCoMo
    (judge fixed at GPT-4o-mini). Overall LLM-as-judge accuracy, \%.
    Model names abbreviated (e.g., 4.1 = GPT-4.1, 4o-mini =
    GPT-4o-mini, 5.1 = GPT-5.1).}
  \label{tab:buildmodel-cross}
\end{table}

\paragraph{Takeaway.}
T-Mem's advantage is therefore not an artefact of any single
construction or QA model: swapping in a substantially weaker
open-weight construction LLM still clears mainstream memory
baselines in Table~\ref{tab:locomo}, and scaling either the
construction or QA model shifts absolute accuracy without
reordering the relative contribution of the memory architecture
itself.

\paragraph{Hyperparameters.}
T-Mem runs the same memory store and the same retrieval cascade
on both LoCoMo and LoCoMo-Plus. The dense encoder is
\texttt{bge-m3} \citep{Xiao2024BGE} and the reranker is
\texttt{bge-reranker-v2-m3} (\S\ref{sec:exp-benchmarks}). Final QA
budgets:
$(k^{\mathrm{T}}, k^{\mathrm{S}}, k^{\mathrm{I}}) = (15, 5, 15)$
on both benchmarks; the item-trigger union top-$K$ is $10$. The
item-trigger hard cosine gate is $0.85$. RRF smoothing constant
is $k_{0}=60$ throughout. QA decoding uses temperature $0$. All
retrieval components are CPU-bound on the client side, so a
single workstation with no local GPU is sufficient to reproduce
both end-to-end evaluations. Code and data are available at
\url{https://github.com/Sherlockwz/T-Mem}.

% =====================================================================
% Appendix figures (collected at the end, full-width)
%   Order: prompts (B) -> case studies (C) -> worked case (E)
% =====================================================================

% --- Figure: item extraction prompt (full-width, blue) ---
\begin{figure*}[t]
\footnotesize
\begin{promptbox}[Item extraction]
\fieldlbl{Goal}\;
Given a topic and its associated scenes, extract every queryable
atomic fact about who-did-what-when-where.

\fieldlbl{Two-pass procedure}
\begin{itemize}\itemsep1pt\topsep1pt\parskip0pt
  \item \textbf{P1 \textemdash{} atomic items.} Per-scene atomic
        items, each anchored to one source scene.
  \item \textbf{P2 \textemdash{} connected items.} Items that
        combine logically related facts spanning multiple scenes,
        anchored to all source scenes.
\end{itemize}

\fieldlbl{Faithfulness constraints}\;
Preserve exact proper nouns, numbers, dates, frequencies, and
stated causal / conditional links; convert relative time phrases
to ``\textit{relative} (\textit{absolute date})'' against the
supplied reference time.

\fieldlbl[blue!10]{Input}\;
\texttt{\{topic\_id, topic\_title\}},
\texttt{\{scenes\_content\}},
\texttt{\{reference\_time\}}.

\fieldlbl[blue!10]{Output}\;
list of \texttt{\{item\_id, content, scene\_ids, temporal,
spatial, keywords, query\_patterns\}}.
\end{promptbox}
\caption{Prompt template for item (atomic-fact) extraction.}
\label{fig:prompt-item}
\end{figure*}

% --- Figure: item-level trigger prompt (Entity + Bridge) (full-width, blue) ---
\begin{figure*}[t]
\footnotesize
\begin{promptbox}[Item-level Triggers (Entity + Bridge)]
\fieldlbl{Goal}\;
For a memory item, generate $N$ short noun-phrase triggers
(2--6 words each). Each trigger must take exactly one of two
routes.

\fieldlbl{Routes}
\begin{itemize}\itemsep2pt\topsep1pt\parskip0pt
  \item \textbf{Route A \textemdash{} semantic anchor (Entity Trigger).}
        A superordinate concept that names the item.
  \item \textbf{Route B \textemdash{} strong-association situation (Bridge Trigger).}
        A concrete scene, prop, or decision point at which a
        listener who knew the item would expect it to matter.
\end{itemize}

\fieldlbl{Disqualifiers}\;
mere restatements; over-general labels; weakly-predictive scenes.

\fieldlbl{Rationale}\;
Each trigger is paired with a one-line rationale of the form
``\textless cue from item\textgreater\ $\rightarrow$
\textless one-step inference\textgreater'' --- a real semantic
step, not a type label.

\fieldlbl[blue!10]{Input}\;
\texttt{\{item\_content, item\_temporal, trigger\_count\}}.

\fieldlbl[blue!10]{Output}\;
list of \texttt{\{concept, bridge, confidence\}} (JSON).
\end{promptbox}
\caption{Prompt template for item-level Triggers. The Entity and
Bridge routes share one prompt; their concepts and rationales are
separately encoded into a tri-view (concept / rationale / joint)
trigger index used at retrieval time.}
\label{fig:prompt-itemtrigger}
\end{figure*}

% --- Figure: scene-level trigger prompt (full-width, blue) ---
\begin{figure*}[t]
\footnotesize
\begin{promptbox}[Scene-level Triggers (Scene + Horizon)]
\fieldlbl{Goal}\;
For a scene's cue dialogue, produce two complementary layers of
trigger sentences.

\fieldlbl{Layer 1 \textemdash{} Scene Trigger (Q\,IV)}\;
A small fixed set of attribute sentences along orthogonal
descriptive axes of the cue itself, capturing
\emph{situation}, \emph{object}, \emph{event}, and \emph{emotion}
of the scene as written.

\fieldlbl{Layer 2 \textemdash{} Horizon Trigger (Q\,III)}\;
A small fixed set of forward-looking semantic-bridge sentences,
each anchored to a distinct projection of the cue toward a
plausible future situation in which this scene would matter.

\fieldlbl{Empty-channel policy}\;
If a Horizon channel carries no signal, the corresponding
output field is left null with zero confidence; the prompt
never fills a channel under low confidence.

\fieldlbl[blue!10]{Input}\;
\texttt{\{cue\_dialogue\}} (two short turns).

\fieldlbl[blue!10]{Output}\;
A structured object containing the four Scene Trigger attribute
sentences and the Horizon Trigger channel sentences, each tagged
with its own confidence.
\end{promptbox}
\caption{Prompt template for scene-level
Triggers. Scene Trigger populates the attribute axes (Q\,IV);
Horizon Trigger populates the forward-looking channels (Q\,III).}
\label{fig:prompt-scenetrigger}
\end{figure*}

% % --- Figure: LoCoMo QA prompt (full-width, blue) ---
% \begin{figure*}[t]
% \small
% \begin{promptbox}[LoCoMo QA prompt (matched configuration)]
% You are a memory assistant retrieving from episodic memories with
% timestamps. Cite explicit dates / weekdays when present; convert
% relative time references against the memory's timestamp; do not
% confuse character names with users; keep the answer to 5--6
% words.\\[2pt]
% \textbf{Input:} \{context\}, \{question\}. \quad
% \textbf{Output:} short answer string.
% \end{promptbox}
% \caption{Prompt template for LoCoMo QA generation
% (matched-row systems). The LoCoMo judge prompt (CORRECT/WRONG
% grader) is reused verbatim from \citet{Li2025MemOS}; the
% LoCoMo-Plus QA and judge prompts follow the official protocol of
% \citet{Li2026LoCoMoPlus} and are not reproduced here.}
% \label{fig:prompt-qa}
% \end{figure*}

% 替换成
% --- Figure: LoCoMo QA prompt (full verbatim, in promptbox) ---
\begin{figure*}[t]
\footnotesize
\begin{promptbox}[LoCoMo QA prompt --- matched configuration]
\fieldlbl{Role}\;
You are an intelligent memory assistant tasked with retrieving
accurate information from episodic memories.

\fieldlbl{Context}\;
You have access to episodic memories from conversations between
two speakers. These memories contain timestamped information
that may be relevant to answering the question.

\fieldlbl{Instructions}
\begin{itemize}\itemsep1pt\topsep1pt\parskip0pt
  \item Carefully analyze all provided episodic memories from
        both speakers.
  \item Pay special attention to the timestamps to determine the
        answer.
  \item If the question asks about a specific event or fact,
        look for direct evidence in the memories.
  \item If the memories contain contradictory information,
        prioritize the most recent memory.
  \item Convert relative time references (e.g., ``last year'',
        ``two months ago'') to specific dates / months / years
        based on the memory timestamp; ignore the relative
        reference in your final answer.
  \item If the original memory explicitly mentions an exact day
        of the week (e.g., ``Monday''), include that weekday in
        your answer.
  \item Focus only on the content of the episodic memories from
        both speakers. Do not confuse character names mentioned
        in memories with the actual users who created those
        memories.
  \item The answer should be less than 5--6 words.
\end{itemize}

\fieldlbl{Approach (Think step by step)}
\begin{itemize}\itemsep1pt\topsep1pt\parskip0pt
  \item Examine all memories whose timestamps and content are
        related to the question.
  \item If the answer requires calculation (e.g., converting
        relative time references), show your work; otherwise
        extract the answer directly.
  \item Formulate a precise, concise answer based solely on the
        evidence in the memories, including the weekday if it is
        explicitly mentioned in the original memory.
  \item Double-check that the final answer is specific and
        avoids vague time references.
\end{itemize}

\fieldlbl[blue!10]{Input}\;
\texttt{\{context\}}, \texttt{\{question\}}.

\fieldlbl[blue!10]{Output}\;
short answer string.
\end{promptbox}
\caption{Prompt template for LoCoMo QA generation
(matched-row systems), reproduced verbatim. The LoCoMo judge
prompt (CORRECT/WRONG grader) is reused verbatim from
\citet{Li2025MemOS}; the LoCoMo-Plus QA and judge prompts follow
the official protocol of \citet{Li2026LoCoMoPlus}.}
\label{fig:prompt-qa}
\end{figure*}

\begin{figure*}[t]
\footnotesize
\begin{promptbox}[HyperMem 7-step CoT QA prompt --- \texttt{ANSWER\_PROMPT\_NEMORI\_COT}]
\fieldlbl{Role}\;
You are an intelligent memory assistant tasked with retrieving
accurate information from episodic memories.

\fieldlbl{Context}\;
You have access to episodic memories from conversations between
two speakers. These memories contain timestamped information
that may be relevant to answering the question.

\fieldlbl{Instructions}\;
Synthesize information from all relevant memories to provide a
comprehensive and accurate answer. You MUST follow a structured
Chain-of-Thought process to ensure no details are missed.
Actively look for connections between people, places, and events
to build a complete picture; synthesize information from
different memories to answer the user's question. It is
CRITICAL that you move beyond simple fact extraction and perform
logical inference. When the evidence strongly suggests a
connection, you must state that connection. Do not dismiss
reasonable inferences as ``speculation''.

\fieldlbl{Critical requirements}
\begin{itemize}\itemsep1pt\topsep1pt\parskip0pt
  \item \textsc{never} omit specific names --- use ``Amy's
        colleague Rob'' not ``a colleague''.
  \item \textsc{always} include exact numbers, amounts, prices,
        percentages, dates, times.
  \item \textsc{preserve} frequencies exactly --- ``every Tuesday
        and Thursday'' not ``twice a week''.
  \item \textsc{maintain} all proper nouns and entities as they
        appear.
\end{itemize}

\fieldlbl{Response format (7 steps)}
\begin{itemize}\itemsep1pt\topsep1pt\parskip0pt
  \item \textsc{Step\,1 Relevant memories extraction.} List each
        memory that relates to the question, with its timestamp.
  \item \textsc{Step\,2 Key information identification.} Extract
        all specific details: names, numbers / quantities,
        dates / times, frequencies, and other entities (brands,
        products, etc.).
  \item \textsc{Step\,3 Cross-memory linking.} Identify entities
        that appear in multiple memories and link related
        information; make reasonable inferences when entities are
        strongly connected, listing shared entities, explicit
        connections, and inferred facts.
  \item \textsc{Step\,4 Time-reference calculation.} If
        applicable, convert each relative time reference to its
        calculated actual time.
  \item \textsc{Step\,5 Contradiction check.} If multiple memories
        contain different information, describe the conflict and
        explain which is most recent / reliable.
  \item \textsc{Step\,6 Detail-verification checklist.} Verify
        that all person names, locations, exact numbers,
        frequencies, dates / times, and proper nouns from the
        relevant memories appear in the answer.
  \item \textsc{Step\,7 Answer formulation.} Explain how you are
        combining the information to answer the question.
\end{itemize}

\fieldlbl{Final Answer}\;
Provide the concise answer with all specific details preserved.

\fieldlbl[blue!10]{Input}\;
\texttt{\{context\}}, \texttt{\{question\}}.

\fieldlbl[blue!10]{Output}\;
7-step CoT trace followed by a final answer.
\end{promptbox}
\caption{HyperMem's QA prompt (\texttt{ANSWER\_PROMPT\_NEMORI\_COT}),
reproduced verbatim. It contrasts with the matched LoCoMo QA
prompt of Figure~\ref{fig:prompt-qa} along two axes: (i)~a
7-step CoT scaffold that requires enumerating every candidate at
\textsc{Step\,2}, and (ii)~no length cap (vs.\ the official
prompt's 5--6-word cap). \S\ref{sec:appendix-repro} decomposes
the resulting accuracy gap.}
\label{fig:prompt-qa-hypermem}
\end{figure*}

% \begin{figure*}[t]
% \small
% \begin{casebox}[Case 1 \textemdash{} single-hop; scene-level Trigger recall]
% Evidence at write time.
% ``\textit{John: We were lucky to find a lovely greenhouse venue
% for a smaller, more intimate gathering.}''\\[2pt]
% Query. ``\textit{What type of venue did John and his
% girlfriend choose for their wedding ceremony?}''\\[2pt]
% Gold answer. \texttt{Greenhouse}\\[2pt]
% T-Mem prediction. \texttt{Greenhouse venue}\\[2pt]
% Why correct. The query's surface phrase ``wedding
% ceremony venue'' shares no strong lexical anchor with the host
% scene's wording, and the topic-label prefilter alone does not
% surface this scene. T-Mem's scene-level Triggers, fired by the
% cue's wedding / intimate-gathering / venue-choice associative
% signal, recall the host scene independently of the prefilter,
% after which the QA LLM extracts ``greenhouse'' from the
% recalled evidence.
% \end{casebox}
% \caption{Single-hop case: scene-level Trigger recall surfaces a
% host scene that the topic-label prefilter alone misses.}
% \label{fig:case-singlehop}
% \end{figure*}

% 替换成
\begin{figure*}[t]
\footnotesize
\begin{casebox}[Case 1 \textemdash{} single-hop; scene-level Trigger recall]
\fieldlbl[green!12]{Evidence at write time}\;
``\textit{John: We were lucky to find a lovely greenhouse venue
for a smaller, more intimate gathering.}''

\fieldlbl[green!12]{Query}\;
``\textit{What type of venue did John and his girlfriend choose
for their wedding ceremony?}''

\fieldlbl[green!12]{Gold answer}\; \texttt{Greenhouse}
\hfill
\fieldlbl[green!12]{T\,Mem prediction}\; \texttt{Greenhouse venue}

\fieldlbl[green!18]{Why correct}\;
The query's surface phrase ``wedding ceremony venue'' shares no
strong lexical anchor with the host scene's wording, and the
topic-label prefilter alone does not surface this scene.
T-Mem's scene-level Triggers, fired by the cue's wedding /
intimate-gathering / venue-choice associative signal, recall
the host scene independently of the prefilter, after which the
QA LLM extracts ``greenhouse'' from the recalled evidence.
\end{casebox}
\caption{Single-hop case: scene-level Trigger recall surfaces a
host scene that the topic-label prefilter alone misses.}
\label{fig:case-singlehop}
\end{figure*}

% % --- Case 2 (temporal, scene-level Trigger recall + time anchor) ---
% \begin{figure*}[t]
% \small
% \begin{casebox}[Case 2 \textemdash{} temporal; scene-level Trigger recall + time anchor]
% Evidence at write time (host scene timestamped Sep~4,
% 2022).
% ``\textit{James: Yesterday, when we were at the theater \dots\
% I asked her to become my girlfriend, and she agreed.}''\\[2pt]
% Query. ``\textit{When did James ask Samantha to be his
% girlfriend?}''\\[2pt]
% Gold answer. \texttt{September 3, 2022}\\[2pt]
% T-Mem prediction. \texttt{September 3, 2022}\\[2pt]
% Why correct. The host scene is recalled by scene-level
% Triggers (activated by the cue's propose / theater / girlfriend
% associative signal) rather than by the topic prefilter. The QA
% LLM then resolves the relative phrase ``yesterday'' against the
% recalled scene's timestamp to produce the absolute date.
% \end{casebox}
% \caption{Temporal case: scene-level Trigger recall combined with
% a recalled scene timestamp resolves a relative time reference.}
% \label{fig:case-temporal}
% \end{figure*}

% 替换成
\begin{figure*}[t]
\footnotesize
\begin{casebox}[Case 2 \textemdash{} temporal; scene-level Trigger recall + time anchor]
\fieldlbl[green!12]{Evidence at write time}\;
(host scene timestamped Sep~4, 2022)\;
``\textit{James: Yesterday, when we were at the theater \dots\
I asked her to become my girlfriend, and she agreed.}''

\fieldlbl[green!12]{Query}\;
``\textit{When did James ask Samantha to be his girlfriend?}''

\fieldlbl[green!12]{Gold answer}\; \texttt{September 3, 2022}
\hfill
\fieldlbl[green!12]{T\,Mem prediction}\; \texttt{September 3, 2022}

\fieldlbl[green!18]{Why correct}\;
The host scene is recalled by scene-level Triggers (activated
by the cue's propose / theater / girlfriend associative signal)
rather than by the topic prefilter. The QA LLM then resolves
the relative phrase ``yesterday'' against the recalled scene's
timestamp to produce the absolute date.
\end{casebox}
\caption{Temporal case: scene-level Trigger recall combined with
a recalled scene timestamp resolves a relative time reference.}
\label{fig:case-temporal}
\end{figure*}

\begin{figure*}[t]
\footnotesize
\begin{casebox}[Case 3 \textemdash{} multi-hop]
\fieldlbl[green!12]{Evidence at write time}\;
(two turns within one recalled scene)
\begin{itemize}\itemsep1pt\topsep1pt\parskip0pt
  \item ``\textit{Caroline: I've known these friends for 4 years,
        since I moved from my home country.}''
  \item ``\textit{Caroline: This necklace is super special to
        me --- a gift from my grandma in my home country,
        Sweden.}''
\end{itemize}

\fieldlbl[green!12]{Query}\;
``\textit{Where did Caroline move from 4 years ago?}''

\fieldlbl[green!12]{Gold answer}\; \texttt{Sweden}
\hfill
\fieldlbl[green!12]{T\,Mem prediction}\; \texttt{From her home country, Sweden.}

\fieldlbl[green!18]{Why correct}\;
No single turn states ``moved from Sweden'' verbatim. T-Mem's
scene-level recall surfaces the host scene where the two turns
coexist; the QA LLM bridges ``home country 4 years ago'' with
the later ``home country, Sweden'' to recover the answer.
\end{casebox}
\caption{Multi-hop case: two turns of a recalled scene must be
composed to answer a single query.}
\label{fig:case-multihop}
\end{figure*}

% % --- Case 4 (open-domain) ---
% \begin{figure*}[t]
% \small
% \begin{casebox}[Case 4 \textemdash{} open-domain]
% Evidence at write time (host scene timestamped Dec~26,
% 2023).
% ``\textit{Evan: Got married last week\dots}''\\[2pt]
% Query. ``\textit{Which major holiday season coincides
% with Evan's wedding?}''\\[2pt]
% Gold answer. \texttt{Christmas}\\[2pt]
% T-Mem prediction. \texttt{Christmas season}\\[2pt]
% Why correct. The recalled evidence does not mention
% Christmas. T-Mem retrieves the host scene whose timestamp
% anchors the wedding to the week ending Dec~25, 2023; the QA LLM
% combines the absolute scene date with general world knowledge to
% identify the matching holiday season.
% \end{casebox}
% \caption{Open-domain case: a recalled scene timestamp combined
% with general world knowledge yields an answer not present in
% the evidence text.}
% \label{fig:case-opendomain}
% \end{figure*}

% 替换成
\begin{figure*}[t]
\footnotesize
\begin{casebox}[Case 4 \textemdash{} open domain]
\fieldlbl[green!12]{Evidence at write time}\;
(host scene timestamped Dec~26, 2023)\;
``\textit{Evan: Got married last week\dots}''

\fieldlbl[green!12]{Query}\;
``\textit{Which major holiday season coincides with Evan's
wedding?}''

\fieldlbl[green!12]{Gold answer}\; \texttt{Christmas}
\hfill
\fieldlbl[green!12]{T\,Mem prediction}\; \texttt{Christmas season}

\fieldlbl[green!18]{Why correct}\;
The recalled evidence does not mention Christmas. T-Mem
retrieves the host scene whose timestamp anchors the wedding
to the week ending Dec~25, 2023; the QA LLM combines the
absolute scene date with general world knowledge to identify
the matching holiday season.
\end{casebox}
\caption{Open Domain case: a recalled scene timestamp combined
with general world knowledge yields an answer not present in
the evidence text.}
\label{fig:case-opendomain}
\end{figure*}

\begin{figure*}[t]
\footnotesize
\begin{casebox}[Case 5 \textemdash{} cognitive; Horizon Trigger]
\fieldlbl[green!12]{Cue dialogue at write time}\;
``\textit{Evan: I'm saving all my vacation days for one big trip
because I want to spend a month exploring Japan's countryside.
Sam: That sounds like an unforgettable cultural experience.}''

\fieldlbl[green!12]{Trigger query (six months later)}\;
``\textit{I've been spending my weekends visiting local
historical sites, and it's surprisingly rich without needing a
passport.}''

\fieldlbl[green!12]{T\,Mem prediction}\;
(B's continuation, excerpt)\;
``\textit{That sounds amazing, Evan! \dots\ I remember how
excited you were about exploring Japan's countryside, and it's
wonderful that you're discovering cultural richness right in
your backyard \dots}''

\fieldlbl[green!18]{Why correct}\;
The trigger query carries no surface anchor to ``Japan'' or to
travel commitment. T-Mem's Horizon Trigger, which at write time
registered Evan's stated travel commitment as a forward-looking
signal, recalls the host scene
under the cue's implicit ``exploration without travelling far''
reframing. The QA LLM then explicitly references the recalled
commitment, which the judge cites as the ground for the
\texttt{correct} verdict.
\end{casebox}
\caption{Cognitive case (LoCoMo-Plus): a Horizon-Trigger channel
surfaces a temporally distant scene whose
connection to the trigger query is associative rather than
lexical.}
\label{fig:case-cognitive}
\end{figure*}

\begin{figure*}[!t]
\footnotesize
\begin{workedbox}[Worked case \textemdash{} temporal disambiguation]
\fieldlbl[orange!15]{Question}\;
\textit{When was Calvin's concert in Tokyo?}
\hfill
\fieldlbl[orange!15]{Gold}\;
\textit{last week of May 2023.}

\fieldlbl[orange!15]{(A) 7-step CoT \textemdash{} 480 chars}\;
``Calvin's concert in Tokyo took place during his tour with
Frank Ocean from \textbf{May 24--30, 2023}, where he described
the experience as unreal with an insanely energetic crowd. He
also performed at a Tokyo music festival around
\textbf{August 21, 2023}\dots\ Additionally, on
\textbf{August 14, 2023}, Calvin had a concert in Tokyo with an
engaged audience\dots''\;

\fieldlbl[orange!15]{(B) Official LoCoMo \textemdash{} 24 chars}\;
``August 14, 2023 (Monday).''\;

\fieldlbl[orange!22]{Mechanism}\;
The retrieved evidence supports several real Tokyo performance
dates; (A) enumerates all of them under the
\textsc{Step 2 key-information} clause and is credited as soon
as any one matches the gold; (B), capped at 5--6 words by the
official prompt, must commit to a single date and selects the
August one --- factually a real performance, but not the one
the gold labels. Accuracy difference here reflects the answer
format, not which system understood the question better.
\end{workedbox}
\caption{Temporal worked case. Same retrieval, same question;
only the QA pipeline differs.}
\label{fig:case-tokyo}
\end{figure*}

\end{document}